\documentclass[lettersize,journal]{IEEEtran}
\usepackage{amsmath,amsfonts,nccmath}
\usepackage{algorithmic}
\usepackage{algorithm}
\usepackage{array}
\usepackage[caption=false,font=normalsize,labelfont=sf,textfont=sf]{subfig}
\usepackage{textcomp}
\usepackage{stfloats}
\usepackage{url}
\usepackage{verbatim}
\usepackage{graphicx}
\usepackage{cite}
\usepackage{pdfpages}

\usepackage{amssymb}
\usepackage{amsthm}
\usepackage{tikz}
\usepackage{forest}
\usetikzlibrary{trees,positioning,shapes,shadows,arrows.meta}
\usetikzlibrary{mindmap,shadows,backgrounds}
\usepackage{hyperref}
\usepackage{booktabs}
\usepackage{multirow}
\begin{document}

\title{Exploring the Landscape of Machine Unlearning: A Comprehensive Survey and Taxonomy}


\author{Thanveer Shaik,  Xiaohui Tao, Haoran Xie, Lin Li, Xiaofeng Zhu, and Qing Li
\thanks{Thanveer Shaik and Xiaohui Tao are with 
the School of Mathematics, Physics and Computing, University of Southern Queensland, Queensland, Australia (e-mail: Thanveer.Shaik@usq.edu.au, Xiaohui.Tao@usq.edu.au).}
\thanks{Haoran Xie is with the Department of Computing and Decision Sciences, Lingnan University, Tuen Mun, Hong Kong (e-mail: hrxie@ln.edu.hk)}
\thanks{Lin Li is with the School of Computer and Artificial Intelligence, Wuhan University of Technology, China (e-mail: cathylilin@whut.edu.cn)}
\thanks{Xiaofeng Zhu is with the University of Electronic Science and Technology of China (e-mail: seanzhuxf@gmail.com)}
\thanks{Qing Li is with the Department of Computing, Hong Kong Polytechnic University, Hong Kong Special Administrative Region of China (e-mail: qing-prof.li@polyu.edu.hk).}
}



\maketitle

\begin{abstract}

Machine unlearning (MU) is gaining increasing attention due to the need to remove or modify predictions made by machine learning (ML) models. While training models have become more efficient and accurate, the importance of unlearning previously learned information has become increasingly significant in fields such as privacy, security, and fairness. This paper presents a comprehensive survey of MU, covering current state-of-the-art techniques and approaches, including data deletion, perturbation, and model updates. In addition, commonly used metrics and datasets are also presented. The paper also highlights the challenges that need to be addressed, including attack sophistication, standardization, transferability, interpretability, training data, and resource constraints. The contributions of this paper include discussions about the potential benefits of MU and its future directions. Additionally, the paper emphasizes the need for researchers and practitioners to continue exploring and refining unlearning techniques to ensure that ML models can adapt to changing circumstances while maintaining user trust. The importance of unlearning is further highlighted in making Artificial Intelligence (AI) more trustworthy and transparent, especially with the increasing importance of AI in various domains that involve large amounts of personal user data.
\end{abstract}


\section{Introduction}

Machine learning (ML) refers to the process of training an algorithm to make predictions or decisions based on data~\cite{sarker2021machine}. ML has become increasingly important in applications such as health, higher education, and other relevant domains. In healthcare, ML models can be used to predict patient outcomes, identify high-risk patients and personalize treatment plans~\cite{davenport2019potential}. For higher education, ML has been used to improve student outcomes and enhance the learning experience, or even used to analyze student data and predict their online class engagement. 

In ML, an algorithm is trained on a dataset to learn patterns and relationships in the data. Once the algorithm has been trained, it can be used to make predictions on new data. Thus, the goal of ML is to create accurate models that can generalize well onto new data~\cite{janiesch2021machine}. On the other hand, machine unlearning (MU) is the process of removing certain data points or features from a trained ML model without affecting its performance~\cite{pmlr-v130-izzo21a}. MU is a relatively new and challenging field of research that is concerned with developing techniques for removing sensitive or irrelevant data from trained models. The goal of MU is to ensure that trained models are free from biases and sensitive information that could lead to negative outcomes~\cite{zhang2023review}.

MU was first introduced by Cao et al.~\cite{Cao2015}, who recognized the need for a ``forgetting system'' and developed one of the initial unlearning algorithms called \emph{machine unlearning}. This approach efficiently removes data traces by converting learning algorithms into a summation form which can help counter data pollution attacks. The increasing need for regulatory compliance with modern privacy regulations led to the establishment of MU, which involves deleting data not only from storage archives, but also from ML models~\cite{eisenhofer2022verifiable}. Existing studies update the model weights for unlearning using either the whole training data, a subset of training data, or some metadata stored during training~\cite{nguyen2022survey}. Although strict regulatory compliance requires the timely deletion of data, there are instances where data pertaining to the training process may not be available for unlearning purposes. Companies and organizations commonly employ user data to train ML models, but legal frameworks like GDPR, CCPA, and CPPA demand that user data be erased when requested~\cite{sekhari2021remember}. The question is whether merely deleting the data is sufficient, or if the models trained on this data should also be adjusted~\cite{mehta2022deep}. However, straightforward techniques like retraining models from scratch or check-pointing can be computationally costly and require significant storage resources~\cite{verkijk2021medroberta}. With MU, we can modify models to exclude specific data points more efficiently~\cite{10.1145/3488932.3517406}.  


Various techniques have been suggested for managing user data deletion requests, such as optimization, clustering, and regression methods. Conducting a comprehensive survey of existing literature on managing user data deletion requests can support the identification of gaps and trends in the field, which will guide future research and provide insights for organizations handling such requests. In this study, we aim to address the following research questions.
\begin{enumerate}
    \item What are the most effective techniques for unlearning data from ML models?
    \item How can the impact of unlearning on model performance be measured and evaluated?
    \item What are the challenges in MU, and how can these challenges be addressed?
\end{enumerate}

The contributions of this study are as follows:
\begin{itemize}
    \item A comprehensive and up-to-date taxonomy about the emerging field of MU, including an explanation of its importance and potential applications.
    \item A detailed taxonomy of the various techniques and approaches that have been developed for unlearning data from ML models, such as data deletion, data perturbation, and model update techniques.
    \item A discussion of different evaluation methods for assessing the effectiveness of MU techniques, such as measuring the degree of forgetting or their impact on model performance.
    \item A taxonomy of several key challenges in the field of MU, including attack sophistication, standardization, transferability, interpretability, training data, and resource constraints.
    \item Finally, a discussion of the potential benefits of MU and its future directions in natural language processing (NLP), computer vision, and recommender systems.
\end{itemize}

The remainder of the paper is organized as follows. Section~\ref{overview} outlines the aims and objectives of MU. In Section~\ref{techniques}, we delve into data deletion, data perturbation, and model update techniques in greater depth. Section~\ref{evaluation} details the evaluation metrics of MU, while Section~\ref{challenges} discusses the challenges associated with the field and proposes potential solutions. In Section~\ref{future}, we explore the future directions of MU in NLP, computer vision, and recommender systems. Finally, Section~\ref{conclusion} concludes the paper.

\section{Overview of Machine Unlearning}\label{overview}


The concept of the ``right to be forgotten'' refers to the ability to have personal information removed from online search results and directories in certain situations~\cite{guimaraes2023preserving}. However, there is no consensus on its definition, or whether it should be classified as a human right. Nevertheless, various institutions and governments, such as those in Argentina, the European Union (EU), and the Philippines, are beginning to propose regulations around this issue~\footnote{https://link.library.eui.eu/portal/The-Right-To-Be-Forgotten--A-Comparative-Study/tw0VHCyGcDc/}. Information and events from an individual's past can continue to carry a stigma and lead to negative consequences, even after a considerable amount of time has passed. For example, in July 2018, Disney fired writer and director James Gunn, who was credited for much of the studio's success with films such as ``Guardians of the Galaxy'', after old tweets resurfaced containing dark humor about pedophilia and rape. Fans and actors rallied to Gunn's defense, with an open letter calling for his reinstatement and petitions to rehire him~\cite{fortner2020decision}. However, deleting what a person has posted on social media platforms such as Facebook and Instagram may not entirely remove the data from the Internet. Although Facebook launched a tool called “Off-Facebook Activity” to help users delete data that third-party apps and websites share with Facebook, it only de-links the data from the user. In 2014, a Spanish court ruled in favor of a man who requested that Google remove certain information about him from its search results~\cite{globocnik2020right}. The court found that the information was no longer relevant or adequate, as the debt had been paid a long time before. The EU court also ruled that Google needed to remove the search results.

The ``right to be forgotten'' is used to describe an individual's right to request that their personal information be removed from the Internet, particularly search engine results, in certain cases~\cite{villaronga2018humans}. Supporters of this right argue that it is necessary to protect individuals from having past mistakes or personal information used against them, such as in cases of revenge porn, petty crimes, or unpaid debts. However, critics of this right claim that it infringes upon freedom of expression and the right to criticize. The EU has tried to address these concerns by striking a balance between the right to privacy and freedom of expression~\cite{hoofnagle2019european}. The issue is further complicated by the use of ML, which can collect and analyze vast amounts of data indefinitely. This data can then be used in applications such as insurance, medical, and loan evaluations, leading to potential harm and amplifying existing biases. As such, it is important to consider the ethical implications of ML models and data collection in these contexts.
We generate a definition of MU based on a comprehensive review of existing research literature, including the studies in~\cite{Cao2015,Bourtoule2021,Aldaghri2021,nguyen2022survey,brophy2021machine,gupta2021adaptive,tarun2023fast}: 

\textbf{General Definition:} \textit{Machine unlearning is a concept that refers to the process of removing or ``forgetting'' previously learned information from a machine learning model. In essence, it is the opposite of machine learning; while machine learning is all about training models to recognize patterns and make predictions based on that data, machine unlearning aims to undo or reverse that process, by removing previously learned patterns or predictions that are no longer relevant or accurate.}

MU is an emerging field within the realm of artificial intelligence (AI) that seeks to remove specific data points from a model without compromising its performance. This technique, also known as selective amnesia~\cite{greengard2022can}, has a variety of potential applications, including enabling individuals to exercise their ``right to be forgotten'' and preventing AI models from inadvertently leaking sensitive information. MU can also help combat data poisoning and adversarial attacks. \textbf{Through its application, MU has the following objectives}:

\begin{itemize}
    \item \textit{To address privacy concerns in ML by eliminating sensitive or personal data from the model without significantly reducing its performance.} It is different from ML, which focuses on training models to predict outcomes based on input data. The works cited in this context include research on novel techniques for privacy-preserving ML, statistical methods for data protection, and adaptive algorithms that adjust to changing data privacy requirements~\cite{graves2021novel,wasserman2010statistical,phan2017adaptive,chen2022proactively,song2021understanding,kuppa2021towards,santos2021consent,saquella2020personal,hinds2020wouldn,Aljeraisy2021,schnitzler2023analyzing}.
    \item \textit{To improve the accuracy and fairness of ML models by removing biases or correcting errors that may have been introduced during the learning process.} This is typically done by analyzing the model's performance on various metrics and identifying areas where improvement is needed. These works cover research on mitigating bias in machine learning models and techniques for improving fairness in algorithmic decision-making~\cite{li2022discover,mehta2019high,madva2017biased,dinsdale2022fedharmony,sarkar2021trapdoor,ashraf2018learning,he2019unlearn,bevan2022skin,risser2022survey,du2022towards,han2023everybody}.
    \item \textit{To improve the performance of ML models over time by allowing them to adapt to changing data and circumstances.} By unlearning outdated or irrelevant information, ML models can become more accurate, efficient, and adaptable to new situations. The references cited in this context include studies on transfer learning, which involves applying knowledge from previously learned tasks to new problems~\cite{xu2020transfer,autz2022pitfalls,riemer2018learning,casella2022transfer,Fu2020,Bu2020,NIPS2014_375c7134,zhang2022survey}.
\end{itemize}

Despite the significant investment that companies make in training and deploying large AI models, regulators in both the EU and the United States are cautioning that models trained on sensitive data may need to be removed. In a report focused on AI frameworks, the UK government explained that ML models may be subject to data deletion under the General Data Protection Regulation (GDPR). For instance, Paravision was recently found to have collected millions of facial photos inappropriately and was required by the US Federal Trade Commission to delete both the data and any trained models that relied on it~\footnote{https://www.wired.com/story/startup-nix-algorithms-ill-gotten-facial-data/}. The most straightforward strategy for removing a data point from training data and updating the model is to conduct retraining~\cite{yu2017compressing}. Unfortunately, this procedure incurs considerable costs, as exemplified by OpenAI's reported expenses of up to 20 million dollars to train GPT-3~\cite{zhao2023survey}. Hence, there is a need for more cost-effective and efficient methods to address data point removal in ML models.


The challenge is to balance privacy and the right to expression to prevent the right to be forgotten from becoming a form of censorship~\cite{kulathoor2022right}. Balancing privacy and the right to expression is crucial in implementing the right to be forgotten without this process being misused. The emergence of new technologies, such as blockchain, presents new challenges in maintaining this balance. Furthermore, the increasing public sensitivity toward data privacy has prompted many companies to prioritize user privacy. For example, Google recently announced an expanded policy for US citizens to remove personal data from search results\footnote{https://techcrunch.com/2022/09/28/google-rolls-out-tool-to-request-removal-of-personal-info-from-search-results-will-later-add-proactive-alerts/}. However, when data points are eliminated, the AI models trained on them need to be appropriately cleaned up to avoid perpetuating biased or sensitive information. While MU is a complex challenge, various approaches are being tested and developed to address this issue. As regulations on data privacy increase, MU is expected to play a critical role in ensuring that AI models are transparent and ethical.

\section{Techniques and Approaches}\label{techniques}
\begin{figure*}
    \centering
    \includegraphics[scale=0.7]{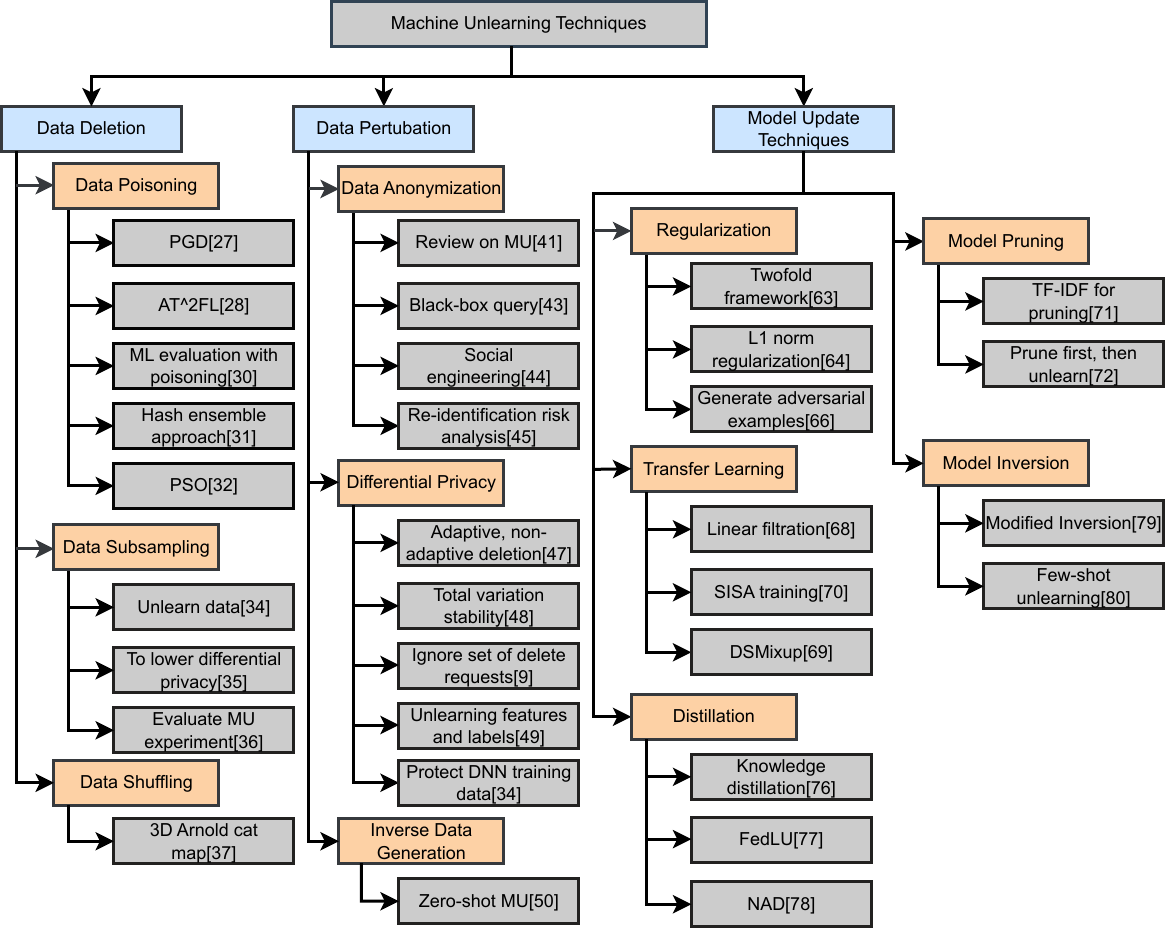}
    \caption{Machine Unlearning - Taxonomy}
    \label{fig:techniques}
\end{figure*}

This section categorizes the MU techniques into three groups, namely Data Deletion, Data Perturbation, and Model Update techniques, as illustrated in the taxonomy in Fig~\ref{fig:techniques}. The first research question will also be addressed in this section.

\subsection{Data Deletion}
In this subsection, we define the data deletion techniques such as data poisoning, data subsampling, and data shuffling.

\subsubsection{\textbf{Data poisoning}}

Data poisoning is a technique used in MU to intentionally introduce incorrect or misleading data into the training dataset. The goal of data poisoning is to degrade the accuracy of the ML model, often with malicious intent. This technique is often used in attacks on privacy-preserving systems or to manipulate the results of automated decision-making processes.

Suppose we have a dataset $D = {(x_{1},y_{1}), (x_{2},y_{2}), ..., (x_{n},y_{n})}$ and a malicious adversary wants to inject a backdoor into the model by modifying a fraction of the training data. The attacker adds a poison data point $(x', y')$ to the training data, with the goal of making the model predict a specific target label $y\_target$ instead of the true label $y'$. The poisoned dataset can be written as:

\begin{equation}
    D' = {(x_1,y_1), (x_2,y_2), ..., (x_i,y_{target}), ..., (x_n,y_n)}
\end{equation}
\text{where} $(x_i,y_{target})$ \text{ is the poisoned data point.}

To minimize the loss function $L(\theta; D')$ but subject to the constraint that the model accuracy on the original training data D, denoted by $Acc(\theta; D)$, does not fall below a predefined threshold of $Acc0$, we can write:
\begin{equation}
 D = (x1, y1), (x2, y2), ..., (xn, yn)
\end{equation}
$minimize L(\theta; D’)$ subject to $Acc(\theta; D) \geq Acc0$

The process of data poisoning involves an attacker identifying vulnerabilities in the data collection process and then submitting maliciously crafted data into the system. The malicious data is often designed to look like legitimate data to better evade detection. Once the malicious data is introduced, the ML model can become biased or produce incorrect results.

A projected gradient descent (PGD) solution is formulated for the data poisoning problem by Marchant et al.~\cite{marchant2022hard}. Their article discusses the challenge of complying with data protection regulations, such as the right to erasure, when it comes to trained ML models. They identified a new vulnerability in ML systems, namely ``poisoning attacks'' that slow down unlearning. Their suite of experiments explores the effects of these attacks in various settings and highlights the risks of deploying approximate unlearning algorithms with data-dependent runtimes. Marchant et al.~\cite{marchant2022hard} call into question the extent to which unlearning improves performance over full retraining, showing that data poisoning can harm computation beyond accuracy, similar to conventional denial-of-service attacks. Sun et al.~\cite{Sun2022} discussed the threat of how attackers can utilize federated learning to launch data poisoning attacks on different nodes. These authors demonstrated a data poisoning attack on Federated Multitask Learning~\cite{smith2017federated}, by formulating an optimal strategy as a general bi-level optimization problem. They also defined three attacks: a direct attack, an indirect attack, and a hybrid attack. In a direct attack, all the target nodes are directly injected with poisoned data while training, whereas in the indirect mode of attack, the attackers target related devices due to communication protocols. In hybrid attack mode, the attackers adopt both direct and indirect attacks. To overcome these attacks, the authors proposed an attack on federated learning ($AT^2FL$) framework wherein implicit gradients of poisoned data can be computed inside source attacking nodes.

Data poisoning can be used to manipulate the training with adversarial attacks, such as random label flipping and distance-based label flipping attacks. In their study, Yerlikaya et al.\cite{Yerlikaya2022} did empirical experiments to check the performances of six ML algorithms under the two adversarial attacks. The authors used spam, botnet, malware, and cancer detection datasets to evaluate the algorithms by launching adversarial attacks on them. The results showed that algorithm behavior depends on the type of dataset. Poisoning attacks typically involve maliciously altering the training dataset to decrease classification accuracy or misclassifying specific inputs when the model is deployed. Thus, hash-based ensemble approaches have been proposed as a solution to counteract poisoning attacks, but their effectiveness in different scenarios such as tabular datasets and ensemble-based ML algorithms (e.g. Random Forests) has not been fully evaluated. The robustness of a hash-based ensemble approach against data poisoning in a tabular dataset was evaluated by Anisetti et al.~\cite{Anisetti2022} using a Random Forest (RF) algorithm as a worst-case scenario. Their results showed that even small ensembles can protect against poisoning and that plain RFs are highly sensitive to label flipping, but almost insensitive to other perturbations. In data poisoning circumstances, selecting the hyperparameters for deep learning (DL) models is critical to maintaining or enhancing the performance metrics. Maabreh et al.~\cite{Maabreh2022} proposed developing DL models that are optimized using the nature-inspired algorithm, particle swarm optimizer (PSO)~\cite{Bonyadi2017}, while some of the training data samples are fake i.e. poisoned data. The results showed that an increase in the poisoning rate decreases all the performance metrics, such as accuracy, recall, precision, and F1-score. PSO can recommend different values for important parameters and improve model performance, even with a high poisoning rate. However, caution should be taken when using PSO, as it may temporarily hide the existence of fake samples and fail when there is a significant concentration of poison in the dataset.  

There are several approaches to defend against data poisoning attacks, including robust training methods that can identify and remove malicious data, as well as techniques that can detect changes in the data distribution. However, sophisticated attackers can also bypass these techniques, so ongoing research is needed to develop more effective defenses.

\begin{table*}[]
\large
\centering
\caption{Data deletion techniques for Machine Unlearning}
\label{tab:data_deletion}
\resizebox{\textwidth}{!}{%
\begin{tabular}{@{}llll@{}}
\toprule
\multicolumn{1}{c}{\textbf{Technique}} &
  \multicolumn{1}{c}{\textbf{Reference}} &
  \multicolumn{1}{c}{\textbf{Problem}} &
  \multicolumn{1}{c}{\textbf{Proposed Framework}} \\ \midrule
\multicolumn{1}{l}{\multirow{5}{*}{Data Poisoning}} &
  \multicolumn{1}{l}{Marchant et al.~\cite{marchant2022hard}} &
  \multicolumn{1}{l}{Complying with data protection regulations} &
  \multicolumn{1}{l}{Projected gradient descent (PGD)} \\ \cmidrule(l){2-4} 
\multicolumn{1}{l}{} &
  \multicolumn{1}{l}{Sun et al.~\cite{Sun2022}} &
  \multicolumn{1}{l}{Data poisoning attack on Federated Learning} &
  \multicolumn{1}{l}{Attack on Federated Learning ($AT^2FL$) framework.} \\ \cmidrule(l){2-4} 
\multicolumn{1}{l}{} &
  \multicolumn{1}{l}{Yerlikaya et al.~\cite{Yerlikaya2022}} &
  \multicolumn{1}{l}{Adversarial attacks on machine learning algorithms} &
  \multicolumn{1}{l}{\begin{tabular}[c]{@{}l@{}}Evaluation of machine learning algorithm performances \\ in attacks using data poisoning.\end{tabular}} \\ \cmidrule(l){2-4} 
\multicolumn{1}{l}{} &
  \multicolumn{1}{l}{Anisetti et al.~\cite{Anisetti2022}} &
  \multicolumn{1}{l}{\begin{tabular}[c]{@{}l@{}}Poisoning attacks to decrease classification accuracy \\ or misclassify specific inputs.\end{tabular}} &
  \multicolumn{1}{l}{Hash-based ensemble approaches.} \\ \cmidrule(l){2-4} 
\multicolumn{1}{l}{} &
  \multicolumn{1}{l}{Maabreh et al.~\cite{Maabreh2022}} &
  \multicolumn{1}{l}{Selecting ideal model learning parameters in data poisoning} &
  \multicolumn{1}{l}{Particle swarm optimization (PSO).} \\ \midrule
\multicolumn{1}{l}{\multirow{3}{*}{Data Subsampling}} &
  \multicolumn{1}{l}{Thudi et al.~\cite{thudi2022bounding}} &
  \multicolumn{1}{l}{Consequences for the problem of unlearning} &
  \multicolumn{1}{l}{Dataset sub-sampling to unlearn data.} \\ \cmidrule(l){2-4} 
\multicolumn{1}{l}{} &
  \multicolumn{1}{l}{Koskela et al.~\cite{koskela2023practical}} &
  \multicolumn{1}{l}{Increase in differential privacy} &
  \multicolumn{1}{l}{Data subsampling strategy to lower differential privacy.} \\ \cmidrule(l){2-4} 
\multicolumn{1}{l}{} &
  \multicolumn{1}{l}{Fan et al.\cite{fan2022fast}} &
  \multicolumn{1}{l}{Machine unlearning evaluation} &
  \multicolumn{1}{l}{Sampling strategy to evaluate a MU experiment.} \\ \midrule
Data Shuffling &
  Musanna et al.~\cite{Musanna2018} &
  Image encryption &
  A 3D Arnold cat map and Fisher-Yates shuffling algorithm. \\ \bottomrule
\end{tabular}}
\end{table*}
\subsubsection{\textbf{Data Subsampling}}

Data subsampling is a technique used in MU to reduce the amount of data used in the model training process. In this technique, a subset of the original data is randomly selected, and only that subset is used to train the model. This technique can be useful in cases where the original dataset is very large and the computational resources required for training the model on the full dataset are prohibitive.

Let $X = {x_1, x_2, ..., x_n}$ be the original training dataset with corresponding labels $Y = {y_1, y_2, ..., y_n}$. We randomly select a subset $S$ of size $m < n$, such that $S = {s_1, s_2, ..., s_m}$. We remove the selected subset from $X$ and $Y$ to create new training sets $X'$ and $Y'$:

\begin{equation}
\begin{aligned}
X' = {x_1, x_2, ..., x_n} - S \\
Y' = {y_1, y_2, ..., y_n} - S  
\end{aligned}
\end{equation}

The MU process is then repeated with the new training set $(X', Y')$ to update the model. This process can be repeated multiple times with different subsets $S$ to further decrease the model's dependence on the original training data.

Thudi et al.~\cite{thudi2022bounding} note how subsampling data before training greatly contributes to reducing the risk of membership inference (MI) attacks, and has further benefits for benchmarks commonly studied in academic literature. The theoretical bounds on all MI attacks have consequences for the problem of unlearning, where MI accuracy on the point to be unlearned is a measure of how well a model has unlearned it. Similarly, Koskela et al.~\cite{koskela2023practical} proposed two strategies: (1) make the subset of data used for tuning smaller than the data used for training for subsequent models, allowing for the extrapolation of hyperparameter values; and (2) train the subsequent models with datasets of the same size as the tuning set. This novel subsampling strategy is used to lower differential privacy and also compute the cost hyperparameter tuning of differential privacy. A sampling strategy by Fan et al. \cite{fan2022fast} to evaluate unlearning belief values for each sample used a $k$-fold cross-validation (with $k=4$) for an MU experiment. In this method, the authors trained the Naive Bayes model three times and predicted the probability of each sample belonging to its labeled class. After averaging the predicted probability of all the mislabeled samples, they estimated their average unlearning belief value using a threshold of $\epsilon = 0.9$.

\subsubsection{\textbf{Data Shuffling}}
Data shuffling is a technique used in MU to obscure sensitive information in a dataset by changing the order of the data points. The objective is to make it harder for an adversary to re-identify individuals in the dataset. In data shuffling, the rows or records of a dataset are randomly rearranged so that the original order is lost, while the statistical properties of the data remain unchanged~\cite{Elmahdy2018}.

Let $X$ = ${(x_1, y_1), (x_2, y_2), ..., (x_n, y_n)}$ be the original labeled dataset, where $x_i$ is the $i$th input feature vector and $y_i$ is the corresponding label: 
\begin{enumerate}

\item Shuffle the data to create a new dataset $X'$ with the same number of samples:
\begin{equation}
    X' = { (x'_1, y'_1), (x'_2, y'_2), ..., (x'_n, y'_n) }
\end{equation}
where $(x'_i, y'_i)$ is a random permutation of $(x_i, y_i)$ for all $i$.

\item Split $X'$ into a training set $X_{tr}$ and validation set $X_{val}$:
\begin{equation}
    X_{tr} = { (x'_1, y'_1), (x'_2, y'_2), ..., (x'_{n_{tr}}, y'_{n_{tr}}) }
\end{equation}

\begin{equation}
\begin{split}
    X_{val} = \{ (x'{n_{tr}+1}, y'{n_{tr}+1}), (x'_{n_{tr}+2}, y'_{n_{tr}+2}),\\ ..., (x'_{n}, y'_{n}) \}
\end{split}
\end{equation}

where $n_{tr}$ is the number of samples in the training set and $n_{val}$ is the number of samples in the validation set.

\item Train the ML model using $X_{tr}$ and validate it using $X_{val}$.

\item Repeat steps 1-3 multiple times to obtain a set of models with different initializations and random shuffles.
     
\end{enumerate}

By shuffling the data, we can reduce the effect of any biases or patterns in the original dataset, and obtain more robust and generalizable models. Data shuffling can be performed using various algorithms, such as Fisher-Yates shuffle~\cite{fisher1963statistical} to randomly swaps data points, or Block shuffling~\cite{liu2022block}, which shuffles entire blocks of data while preserving the order within each block. 

An image encryption scheme using a 3-Dimensional (3-D) Arnold cat map and Fisher-Yates shuffling algorithm are proposed by Musanna et al.~\cite{Musanna2018}. The image is divided into slices and shuffled using the 3-D chaotic map. A fractional order system is used for diffusion in the pixel intensity values, which creates a strange attractor. Fisher-Yates is used to create a matrix for arranging data points. The proposed scheme is evaluated on various images, showing high security and sensitivity.

Additionally, there are tools and libraries available for data shuffling, such as the Python library Scikit-learn\footnote{https://scikit-learn.org/stable/modules/generated/sklearn.utils.shuffle.html}, which provides a ShuffleSplit function for shuffling datasets. 

A number of research works focused on data deletion techniques for MU are summarised in Tab.~\ref{tab:data_deletion}.

\subsection{Data Perturbation}

In this subsection, we describe the data perturbation techniques such as data poisoning, data subsampling, and data shuffling.


\subsubsection{\textbf{Data anonymization}}


Data anonymization is the process of removing or modifying personally identifiable information (PII) from a dataset to prevent the identification of individuals~\cite{murthy2019comparative}. This technique is commonly used in MU to protect the privacy of individuals whose data was used in an ML model. The process of anonymization can involve techniques such as masking, generalization, and perturbation.

Let $X$ be the original dataset with sensitive attributes and $X'$ be the anonymized dataset that preserves privacy while retaining utility. Anonymization can be formulated as a mapping $f: X \rightarrow X'$ that satisfies some privacy criteria. One common criterion is $k$-anonymity, which requires that for any record in $X'$, there are at least $k-1$ other records in $X'$ that share the same combination of quasi-identifiers (i.e. attributes that can be used to re-identify individuals).

Let $Q$ be a set of quasi-identifiers and $S$ be the set of sensitive attributes. The $k$-anonymity criterion can be expressed as:
\begin{equation}
\begin{split}
    \forall x' \in X', \exists S' \subseteq S, |S'|>0, and  |\{x\in X:(x[Q]=x'[Q]) \\ \land(x[S']=x'[S'])\}|\geq \textit{k}-1
\end{split}
\end{equation}
where $x[Q]$ and $x[S']$ denote the quasi-identifiers and sensitive attributes of record $x$, respectively.

The authors in Jegorova et al.~\cite{Jegorova2022} discussed the differentiation of personal, personal ``sensitive'', and non-personal data in accordance with the GDPR~\cite{Voigt2017}. Personal data, as defined in Article 4(1) of the GDPR, refers to data that directly or indirectly pertains to an identified or identifiable natural person. Sensitive data, as outlined by the GDPR, includes data that reveals racial or ethnic origin, political opinions, religious or philosophical beliefs, trade-union membership, genetic data, biometric and health-related data, and sexual orientation. Such data requires stronger safeguards around processing, storage, transfer, etc. In contrast, non-personal data refers to data that falls outside the scope of personal data and is significant for data-driven research, analysis, and commercial applications.

While methods are available to distinguish personal data from sensitive data in large language models (LLMs), it is common practice to assume that all data provided for research, even if anonymized, falls under the category of sensitive data due to non-consensual use of data in research. Carlini et al.~\cite{carlini2021extracting} discuss how LLMs can memorize and inadvertently disclose individual training examples, despite a prevailing belief that they do not memorize specific examples due to minimal overfitting. The authors proposed a simple and efficient method for extracting verbatim sequences from a language model's training set using black-box query access. They demonstrated the effectiveness of their attack on the GPT-2 model, extracting sensitive information such as names, addresses, email addresses, phone numbers, and fax numbers. The paper provided a quantitative definition of memorization and analyzes different attack configurations and their impact on the types of data extracted. The authors concluded by discussing strategies to mitigate privacy leakage, including differentially-private training and document de-duplication, but caution that no approach is foolproof.


The evaluation of the verification process implemented by organizations in domains such as finance, entertainment, and retail was conducted by Martino et al.~\cite{di2019personal}. The authors attempted to impersonate individuals and request personal data by using social engineering techniques and publicly available information. They found that policies and practices varied significantly across organizations, and for 15 out of 55, they were able to successfully impersonate individuals and gain access to their personal data. The leaked data included sensitive information such as financial transactions, website visits, and physical location history. Their paper also provides suggestions for policy improvements to minimize the risk of personal information leakage to unauthorized parties. The real-world risk of re-identification depends on factors such as the frequency of data points, the size of source datasets, and the availability of public data to support the re-identification process. A novel re-identification risk analysis framework that considers information reasonably available to an attacker was proposed in Xia et al.~\cite{xia2021enabling}. This framework allows organizations to simulate threats under various degrees of completeness in an attacker's knowledge about individuals in a dataset, providing a more realistic assessment of re-identification risks. The framework can be combined with data utility measures to strike a balance between privacy and utility. A case study in their paper using patient records illustrated that the re-identification risk may be lower than suggested by worst-case scenarios, especially when considering the actual capabilities of attackers who rely on external resources.

\begin{table*}[]
\large
\centering
\caption{Data Perturbation Techniques for Machine Unlearning}
\label{tab:data_perturbation}
\resizebox{\textwidth}{!}{%
\begin{tabular}{@{}llll@{}}
\toprule
\multicolumn{1}{c}{Technique} &
  \multicolumn{1}{c}{Reference} &
  \multicolumn{1}{c}{Problem} &
  \multicolumn{1}{c}{Proposed Framework} \\ \midrule
\multicolumn{1}{l}{\multirow{4}{*}{Data Anonymization}} &
  \multicolumn{1}{l}{Jegorova et al.~\cite{Jegorova2022}} &
  \multicolumn{1}{l}{\begin{tabular}[c]{@{}l@{}}Leakage and Privacy of personal, personal sensitive, \\ and non-personal data\end{tabular}} &
  \multicolumn{1}{l}{A review paper} \\ \cmidrule(l){2-4} 
\multicolumn{1}{l}{} &
  \multicolumn{1}{l}{Carlini et al.~\cite{carlini2021extracting}} &
  \multicolumn{1}{l}{Memorize and leak individual training examples in LLMs} &
  \multicolumn{1}{l}{Extracting verbatim sequences from an LLMs training set} \\ \cmidrule(l){2-4} 
\multicolumn{1}{l}{} &
  \multicolumn{1}{l}{Martino et al.~\cite{di2019personal}} &
  \multicolumn{1}{l}{Verifying the identity of the data requester} &
  \multicolumn{1}{l}{Impersonate individuals and request personal data} \\ \cmidrule(l){2-4} 
\multicolumn{1}{l}{} &
  \multicolumn{1}{l}{Xia et al.~\cite{xia2021enabling}} &
  \multicolumn{1}{l}{Privacy and re-identification risks} &
  \multicolumn{1}{l}{Re-Identification risk analysis framework} \\ \midrule
\multicolumn{1}{l}{\multirow{5}{*}{Differential Privacy}} &
  \multicolumn{1}{l}{Gupta et al.~\cite{gupta2021adaptive}} &
  \multicolumn{1}{l}{Deletion sequences} &
  \multicolumn{1}{l}{\begin{tabular}[c]{@{}l@{}}Deletion guarantees adaptive deletion sequences to \\ non-adaptive deletion sequences\end{tabular}} \\ \cmidrule(l){2-4} 
\multicolumn{1}{l}{} &
  \multicolumn{1}{l}{Ullah et al.~\cite{pmlr-v134-ullah21a}} &
  \multicolumn{1}{l}{Convex risk minimization problems} &
  \multicolumn{1}{l}{Notion of total variation (TV) stability} \\ \cmidrule(l){2-4} 
\multicolumn{1}{l}{} &
  \multicolumn{1}{l}{Sekhari et al.~\cite{sekhari2021remember}} &
  \multicolumn{1}{l}{Problem of unlearning data points from a learnt model} &
  \multicolumn{1}{l}{\begin{tabular}[c]{@{}l@{}}Construct an unlearning algorithm that only depends \\ on the learning algorithm\end{tabular}} \\ \cmidrule(l){2-4} 
\multicolumn{1}{l}{} &
  \multicolumn{1}{l}{Warnecke et al.~\cite{warnecke2021machine}} &
  \multicolumn{1}{l}{Reverting larger groups of features and labels} &
  \multicolumn{1}{l}{\begin{tabular}[c]{@{}l@{}}Framework for unlearning features and labels from \\ learning models\end{tabular}} \\ \cmidrule(l){2-4} 
\multicolumn{1}{l}{} &
  \multicolumn{1}{l}{Thudi et al.~\cite{thudi2022bounding}} &
  \multicolumn{1}{l}{Membership Inference attacks} &
  \multicolumn{1}{l}{Differential privacy to protect deep neural network training data} \\ \midrule
Inverse Data Generation &
  Chundawat et al.~\cite{chundawat2022zero} &
  Removing the information of the forgotten data from the model &
  Zero-shot machine unlearning \\ \bottomrule
\end{tabular}}
\end{table*}

\subsubsection{\textbf{Differential Privacy}}
Differential privacy is a technique that can be used in MU to protect sensitive data, while still allowing analysis to be conducted on the data. The goal of differential privacy is to provide strong privacy guarantees while preserving the utility of the data. This is achieved by adding noise to the data in a way that ensures that the original data cannot be easily reconstructed. In the context of MU, differential privacy can be used to protect the data used to train ML models~\cite{sekhari2021remember}. By adding noise to the data, it becomes more difficult for an attacker to reconstruct the original data and use it to re-identify individuals.

Data deletion algorithms aim to remove the influence of deleted data points from trained models in a computationally efficient manner. However, most existing work in the non-convex setting provides guarantees only for deletion sequences that are chosen independently of the published models. To overcome this, Gupta et al.~\cite{gupta2021adaptive} proposed a general reduction technique that uses differential privacy and its link to maximum information to transform guarantees against adaptive deletion sequences into those against non-adaptive deletion sequences. Their approach allows for flexible algorithms that can handle various model classes and training methodologies, providing strong provable deletion guarantees for adaptive deletion sequences. They also presented a practical attack against the existing Split into Shards and Aggregating (SISA) algorithm~\cite{Bourtoule2021} on CIFAR-10~\cite{krizhevsky2009learning}, MNIST~\cite{lecun1998gradient}, and Fashion-MNIST~\cite{xiao2017fashion}, demonstrating how prior work for non-convex models fails against adaptive deletion sequences. Similarly, Ullah et al.~\cite{pmlr-v134-ullah21a} proposed a notion of total variation (TV) stability to overcome the convex risk in data deletion. This approach allows for exact unlearning in general settings. They also developed TV stable learning algorithms and efficient exact unlearning algorithms for smooth convex empirical risk minimization (ERM) problems, including many ML problems. Their method retrains only a fraction of edit requests, thereby maintaining accuracy and satisfying the criteria of exact unlearning.

The authors of \cite{sekhari2021remember} associated MU and differential privacy due to their strong similarity. They proposed a method of utilizing tools from differential privacy for MU, suggesting that an unlearning algorithm can be created that is independent of the set of delete requests and is reliant only on the learning algorithm. The authors further explain that this unlearning algorithm can satisfy specific conditions regarding privacy and performance, including a guarantee similar to differential privacy. Instead of data deletion, Warnecke et al.~\cite{warnecke2021machine} proposed a novel framework for unlearning features and labels from learning models, based on closed-form updates for model parameters, which is significantly faster than instance-based approaches. The authors also introduced certified unlearning strategies for convex loss models, providing theoretical guarantees on the removal of features and labels. Additionally, their empirical analysis demonstrated that unlearning sensitive information is possible even for deep neural networks (DNNs) with non-convex loss functions, with this proposed framework demonstrably effective in case studies on unlearning sensitive features across linear models, unintended memorization in language models, and label poisoning in computer vision.

The use of differential privacy to safeguard DNN training data against MI attacks is discussed in Thudi et al.~\cite{thudi2022bounding}. The authors proposed a tight bound on MI performance for training algorithms that provides $\epsilon$-DP and extends the result to bound MI positive accuracy for ($\epsilon$, $\delta$)-DP. They showed that one can unlearn under this definition by training with a relatively large $\epsilon$-DP, if dataset sub-sampling is used.

\subsubsection{\textbf{Inverse Data Generation (IDG)}}
Inverse data generation is a technique used in MU to generate a new dataset that is similar to the original dataset, but does not contain any sensitive information. The process involves training a generative model to create a new dataset that mimics the original dataset's statistical properties while omitting sensitive data~\cite{li2022making}. Some popular algorithms for inverse data generation include Generative Adversarial Networks (GANs)~\cite{goodfellow2020generative}, Variational Autoencoders (VAEs)~\cite{kingma2019introduction}, and Deep Belief Networks (DBNs)~\cite{mohamed2011acoustic}. These algorithms are trained to generate new data points that are similar to the original dataset without sensitive information.

The problem of zero-shot MU, where no original data samples are available, is introduced by Chundawat et al. \cite{chundawat2022zero}. They proposed two novel solutions based on error-minimizing-maximizing noise and gated knowledge transfer. These methods remove the information of the forgotten data from the model while maintaining model efficacy on the retained data. The zero-shot approach offers protection against model inversion attacks and membership inference attacks. Their approach introduces a new evaluation metric, the Anamnesis Index (AIN), which measures the quality of the unlearning method effectively. Experimental results on benchmark vision datasets show promising results for unlearning in deep learning models.


The research works that have focused on data perturbation techniques for MU are summarized in Tab.~\ref{tab:data_perturbation}.
\subsection{Model Update Techniques}

\subsubsection{\textbf{Regularization}}
In MU, regularization is a technique used to prevent overfitting by adding a penalty term to the loss function during training~\cite{mehta2019high}. Overfitting occurs when a model becomes too complex and starts to memorize the training data rather than learning general patterns that can be applied to new data. Regularization aims to reduce the complexity of the model by adding a penalty term that encourages the model to have small weights or parameters. This penalty term is usually scaled by a hyperparameter that determines the trade-off between the goodness of fit of the model and the complexity of the model. There are several types of regularization techniques used in MU~\cite{Aldaghri2021}, including L1 regularization, L2 regularization, and dropout regularization. L1 regularization adds a penalty term equal to the absolute value of the weights, while L2 regularization adds a penalty term equal to the square of the weights. Dropout regularization randomly drops out some of the neurons during training, forcing the model to learn more robust and generalizable features.

The thesis of Graves~\cite{graves2021novel} concentrated on the manipulation of trained models after the training phase, specifically model repair, which involves modifying a trained model to improve its suitability for specific tasks. Their presented methods aim to perform the repair in an architecture-agnostic way, maintaining model performance and requiring less computation time than training a model from scratch. The post-training model repair can allow repairs to be done for goals that were not previously considered, such as individual fairness or data privacy regulations. The contributions of their thesis are twofold: (1) the development of algorithmic approaches to remove subsets of learning from a trained neural network, and (2) providing a novel detection method for finding proxy features for individual fairness.

The L1 norm regularization as a penalty item is proposed by Ma et al.~\cite{ma2022learn} to prevent ``over-unlearning'', which may decrease the accuracy of the target model. The authors also proposed a novel unlearning method, Forsaken. The optimization problem, including the KL divergence loss function, the penalty coefficient, and the mask gradient, has this item added to it. The L1 norm is chosen over the $L_{p}$ norm as it is smoother and can avoid blocking the convergence of Forsaken. The regularization term can also be used to extract unlearning errors by calculating model weight differences using $L_{2}$ regularization~\cite{thudi2022unrolling}. A new regularization method based on the inductive bias that the model's predictions should be less confident on out-of-distribution inputs is introduced in Setlur et al.~\cite{setlur2022adversarial}. Their proposed method generates adversarial examples using large step sizes and trains the model to have reduced confidence in them. The method improves in-distribution test accuracy and does not require an additional unlabeled dataset. The model is trained to predict a uniform distribution over labels on the self-generated samples using the principle of maximum entropy.

\subsubsection{\textbf{Transfer Learning}}
Transfer learning is a technique in MU that allows the reuse of knowledge from one ML model to another~\cite{casimiro2021self}. The idea is to use pre-trained models as a starting point for a new model, an approach that can save a lot of time and computational resources. In transfer learning, the knowledge gained from one task is applied to another task. It is a common technique in deep learning, where the pre-trained model is a DNN.




The problem of unlearning in ML involves deleting part of the training data from a model to obtain a model that appears as if it has never seen that data. Retraining the model from scratch may be computationally expensive and impractical for real-world models. Previous approaches towards unlearning have been proposed, but they may not be efficient or applicable in certain settings. The linear filtration algorithm proposed by Baumhauer et al.~\cite{baumhauer2022machine} aims to sanitize classification models that predict logits (re-scaled logarithmic probabilities) after class-wide deletion requests. The proposed method involves applying a linear transformation to the model's predictions that can be absorbed into the original classifier. The computation of transformation is computationally efficient and requires only a small number of data points per class. The authors also proposed a weakened ``black-box'' variant of the definition of unlearning, which may be more realistic and practical in practice. They suggest evaluating the quality of an empirical unlearning operation in an adversarial setting by testing how well it prevents certain privacy attacks on ML models.

In a recent study, Bourtoule et al.~\cite{Bourtoule2021} investigated MU using SISA training and found that by limiting the influence of individual data points during training, they were able to speed up the unlearning process. This is particularly beneficial for stateful algorithms like stochastic gradient descent in DNNs that aim to maximize performance. When combined with transfer learning, SISA training led to a 1.36-fold increase in retraining speed for complex tasks such as ImageNet classification, although accuracy was slightly reduced.

The MU strategy called Dynamically Selected Mix-up (DSMixup) was proposed by Zhou et al.~\cite{zhou2022dynamically} to address the issue of unlearning private data from ML models. The method is based on the SISA~\cite{Bourtoule2021} approach, but instead of retraining all affected shards, it dynamically selects shards that need to be retrained and mixes them together using mix-up data augmentation. This approach avoids the overhead of retraining all shards, improves unlearning efficiency, and maintains system stability. Transfer learning is used to solve the coupling problem of shards caused by mix-up, and the expected number of samples that need to be retrained is analyzed. Experimental results on different datasets by Zhou et al. showed that DSMixup outperformed SISA in both unlearning cost and aggregation model performance.

\begin{table*}[]
\large
\centering
\caption{Model Update Techniques for Machine Unlearning}
\label{tab:model_update}
\resizebox{\textwidth}{!}{%
\begin{tabular}{@{}llll@{}}
\toprule
\textbf{Technique} &
  \textbf{Reference} &
  \textbf{Problem} &
  \textbf{Proposed Framework} \\ \midrule
\multicolumn{1}{l}{\multirow{3}{*}{Regularization}} &
  \multicolumn{1}{l}{Graves~\cite{graves2021novel}} &
  \multicolumn{1}{l}{Post-training model manipulation and model repair} &
  \multicolumn{1}{l}{
  \begin{tabular}{l}
Twofold Framework: algorithmic approaches \\
and a novel detection method.\
\end{tabular}} \\ \cmidrule(l){2-4} 
\multicolumn{1}{l}{} &
  \multicolumn{1}{l}{Ma et al.~\cite{ma2022learn}} &
  \multicolumn{1}{l}{Overlearning} &
  \multicolumn{1}{l}{Add L1 norm regularization as a penalty term} \\ \cmidrule(l){2-4} 
\multicolumn{1}{l}{} &
  \multicolumn{1}{l}{Setlur et al.~\cite{setlur2022adversarial}} &
  \multicolumn{1}{l}{Overlearning} &
  \multicolumn{1}{l}{Generates adversarial examples} \\ \midrule
\multicolumn{1}{l}{\multirow{3}{*}{Transfer Learning}} &
  \multicolumn{1}{l}{Baumhauer et al.~\cite{baumhauer2022machine}} &
  \multicolumn{1}{l}{Transferability} &
  \multicolumn{1}{l}{Linear filtration for sanitizing classification models.} \\ \cmidrule(l){2-4} 
\multicolumn{1}{l}{} &
  \multicolumn{1}{l}{Bourtoule et al.~\cite{bourtoule2021machine}} &
  \multicolumn{1}{l}{Transferability} &
  \multicolumn{1}{l}{SISA training} \\ \cmidrule(l){2-4} 
\multicolumn{1}{l}{} &
  \multicolumn{1}{l}{Zhou et al.~\cite{zhou2022dynamically}} &
  \multicolumn{1}{l}{Unlearning private data from machine learning models} &
  \multicolumn{1}{l}{DSMixup} \\ \midrule
\multicolumn{1}{l}{\multirow{2}{*}{Model Pruning}} &
  \multicolumn{1}{l}{Wang et al.~\cite{wang2022federated}} &
  \multicolumn{1}{l}{Removal of specific categories from trained CNN models in FL} &
  \multicolumn{1}{l}{Using TF-IDF, Pruning on the most discriminative channels.} \\ \cmidrule(l){2-4} 
\multicolumn{1}{l}{} &
  \multicolumn{1}{l}{Jia et al.~\cite{jia2023model}} &
  \multicolumn{1}{l}{Removal of specific categories from trained models} &
  \multicolumn{1}{l}{\begin{tabular}{l}
"Prune first, then unlearn" \
Used OMP~\cite{ma2021sanity}, IMP~\cite{frankle2018lottery}, \\ pruning at random initialization before training~\cite{tanaka2020pruning}\
\end{tabular}} \\ \midrule
\multicolumn{1}{l}{\multirow{3}{*}{Distillation}} &
  \multicolumn{1}{l}{Wu et al.~\cite{wu2022federated}} &
  \multicolumn{1}{l}{To remove a designated client's contribution from the global model} &
  \multicolumn{1}{l}{Knowledge distillation technique} \\ \cmidrule(l){2-4} 
\multicolumn{1}{l}{} &
  \multicolumn{1}{l}{Zhu et al.~\cite{zhu2023heterogeneous}} &
  \multicolumn{1}{l}{Heterogeneous knowledge graph embedding learning and unlearning} &
  \multicolumn{1}{l}{FedLU} \\ \cmidrule(l){2-4} 
\multicolumn{1}{l}{} &
  \multicolumn{1}{l}{Li et al.~\cite{li2021neural}} &
  \multicolumn{1}{l}{Erasing backdoor triggers from backdoored DNNs} &
  \multicolumn{1}{l}{Neuron Attention Distillation (NAD)} \\ \midrule
\multicolumn{1}{l}{\multirow{2}{*}{Model inversion}} &
  \multicolumn{1}{l}{Graves et al.~\cite{Graves2021}} &
  \multicolumn{1}{l}{Post-training model manipulation and model repair} &
  \multicolumn{1}{l}{Modified version of the standard model inversion attack} \\ \cmidrule(l){2-4} 
\multicolumn{1}{l}{} &
  Yoon et al.~\cite{yoon2022few} &
  Post-training model manipulation and model repair &
  Few-shot unlearning using model inversion \\ \bottomrule
\end{tabular}}
\end{table*}

\subsubsection{\textbf{Model Pruning}}
Model pruning is a technique used in ML to reduce the size of a trained model by removing unnecessary parameters while maintaining its performance~\cite{zhao2022novel}. In the context of MU, model pruning can be used to remove sensitive or private data from a model without significantly affecting its accuracy. The process of model pruning involves iteratively removing the least important weights or neurons from the model, based on certain criteria such as their magnitude or contribution to the output~\cite{yeom2021pruning}. This can be done using techniques such as weight magnitude pruning, which removes weights with the smallest absolute values, or sensitivity-based pruning, which removes weights that have the least impact on the output.

An ML unlearning method designed to comply with GDPR requests for the removal of specific categories from trained CNN models in federated learning (FL) was proposed by Wang et al.~\cite{wang2022federated}. Using the Term Frequency - Inverse Data Frequency (TF-IDF), the federated server evaluates the relevance score between the channels and categories and builds a pruner to execute pruning on the most discriminative channels of the target category. After the pruning process is complete, each FL device is notified and downloads the pruned model from the federated server. The devices then conduct normal federated training programs with target category-excluded training data to achieve fine-tuning. Overall, this proposed unlearning method is a complementary block for FL in compliance with legal and ethical criteria. A ``prune first, then unlearn'' paradigm for MU was proposed by Jia et al.~\cite{jia2023model}, which emphasizes that unlearning on a sparse model can lead to a smaller unlearning error and enhance the effectiveness of MU. Different pruning methods such as one-shot magnitude pruning (OMP)~\cite{ma2021sanity}, pruning at random initialization before training~\cite{tanaka2020pruning}, and pruning-training simultaneous iterative magnitude pruning (IMP)~\cite{frankle2018lottery} are discussed. The trainer would prioritize a pruning method that depends the least on the forgetting dataset, and ensures lossless generalization when pruning while improving pruning efficiency. 


\subsubsection{\textbf{Distillation}}
Distillation is an MU technique used to reduce the size of large and complex models while preserving their accuracy. The main idea behind distillation is to train a smaller and simpler model (student model) to mimic the behavior of a larger and more complex model (teacher model), by using the teacher model as a source of guidance~\cite{wang2021knowledge}. In distillation, the teacher model is first trained on a large dataset and then used to generate soft targets for the student model. These soft targets are probability distributions over the output space of the teacher model, which provides further information compared to the hard labels used in traditional training. The student model is then trained on the soft targets, with the objective of minimizing the difference between its predicted probabilities and the soft targets provided by the teacher model.

Wu et al.~\cite{wu2022federated} proposed a novel federated unlearning method to remove a designated client's contribution from the global model after the federated training process. To eliminate the attacker's influence and reduce the unlearning cost, the proposed method uses the knowledge distillation technique~\cite{chen2021knowledge} to erase the historical parameter updates from the attacker and recover the damage. The old global model is used as a teacher to train the unlearning model. This approach has several advantages, including no client-side time and energy costs, no network transmission, and better generalization around training points, all leading to improved model robustness and performance. An FL framework called FedLU for heterogeneous knowledge graph (KG) embedding learning and unlearning is proposed by Zhu et al.~\cite{zhu2023heterogeneous}. To address data heterogeneity, mutual knowledge distillation is used to transfer local knowledge to the global level and vice versa. An unlearning method is also presented to erase specific knowledge from local embeddings and propagate it to the global embedding using knowledge distillation. The effectiveness of the framework was validated on three new datasets based on FB15k-237 using varied numbers of clients.

Li et al.~\cite{li2021neural} proposed a defense framework, Neuron Attention Distillation (NAD), to erase backdoor triggers from backdoored DNNs. NAD is a distillation-guided fine-tuning process that uses a teacher network to guide the fine-tuning of a backdoored student network on a small subset of clean training data, so that the intermediate-layer attention of the student network is well-aligned to that of the teacher network. The teacher network can be obtained from the backdoored student network via standard fine-tuning using the same clean subset of data. Their empirical results showed that NAD is more effective in removing the network's attention on the trigger pattern compared to standard fine-tuning or neural pruning methods.


\subsubsection{\textbf{Model inversion}}
Model inversion is a technique used in MU to extract sensitive data from an ML model. The basic idea behind model inversion is to use a trained model to infer sensitive information about the data used to train the model. This is achieved by manipulating the input to the model to generate a specific output, then using this output to learn something about the input. Let's say an ML model has been trained to classify emails as spam or legitimate based on certain features like keywords and senders. Attackers can use model inversion to determine which specific words or phrases are most likely to trigger the spam filter, allowing them to craft emails that evade detection. Model inversion attacks can be mitigated using various techniques, such as differential privacy, regularization, and adversarial training~\cite{he2019model}.

The modified version of the standard model inversion attack was proposed by Graves et al.~\cite{Graves2021}. This attack begins with a feature vector with all features assigned to 0. The vector is then labeled with the label of the target class, and then a forward and backward pass is performed through the model to obtain the gradient of the loss with respect to the feature vector. Each feature is then shifted in the direction of the gradient iteratively, altering the feature vector to become more similar to what the model considers to be an example from that class. The researchers modified the attack by periodically applying image processing every \textit{n} gradient descent step, then started each inversion with a small amount of noise added to each feature, continuing the attack for some set number of iterations even while the change in the loss was small. These modifications allowed the authors to generate inversions on complex convolutional architectures such as Resnet18~\cite{he2016deep}. A framework for few-shot unlearning using model inversion was proposed by Yoon et al.~\cite{yoon2022few}. This approach involves approximating the original data distribution \textit{D} using model inversion, and filtering out an interpolation of the original distribution. The model inversion is achieved by minimizing the entropy of the model's output and using prior knowledge from a generative model and data augmentation. The resulting approximation \textit{D'} is a set of samples that meet certain criteria for consistency and entropy. Their approach also used a binary classifier to identify noisy samples. 

The research works focused on model update techniques for MU are summarized in Tab.~\ref{tab:model_update}.

\begin{table}[]
\large
\centering
\caption{Public Datasets for Machine Unlearning}
\label{tab:datasets}
\resizebox{1.0\columnwidth}{!}{%
\begin{tabular}{@{}llllll@{}}
\toprule
\textbf{Modality} &
  \textbf{Dataset} &
  \textbf{\begin{tabular}[c]{@{}l@{}}No. of \\ Instances\end{tabular}} &
  \textbf{\begin{tabular}[c]{@{}l@{}}No. of \\ Attributes\end{tabular}} &
  \textbf{Task} &
  \textbf{Popularity*} \\ \midrule
\multicolumn{1}{l}{\multirow{6}{*}{Image}} &
  \multicolumn{1}{l}{SVHN~\cite{netzer2011reading}} &
  \multicolumn{1}{l}{600,000} &
  \multicolumn{1}{l}{3072} &
  \multicolumn{1}{l}{Object recognition} &
  \multicolumn{1}{l}{High} \\ \cmidrule(l){2-6} 
\multicolumn{1}{l}{} &
  \multicolumn{1}{l}{CIFAR-100~\cite{krizhevsky2009learning}} &
  \multicolumn{1}{l}{60,000} &
  \multicolumn{1}{l}{3072} &
  \multicolumn{1}{l}{Object recognition} &
  \multicolumn{1}{l}{High} \\ \cmidrule(l){2-6} 
\multicolumn{1}{l}{} &
  \multicolumn{1}{l}{Imagenet~\cite{deng2009imagenet}} &
  \multicolumn{1}{l}{1.2 million} &
  \multicolumn{1}{l}{1,000} &
  \multicolumn{1}{l}{Object recognition} &
  \multicolumn{1}{l}{Medium} \\ \cmidrule(l){2-6} 
\multicolumn{1}{l}{} &
  \multicolumn{1}{l}{Mini-Imagenet~\cite{vinyals2016matching}} &
  \multicolumn{1}{l}{100,000} &
  \multicolumn{1}{l}{784} &
  \multicolumn{1}{l}{Object recognition} &
  \multicolumn{1}{l}{Low} \\ \cmidrule(l){2-6} 
\multicolumn{1}{l}{} &
  \multicolumn{1}{l}{LSUN~\cite{yu15lsun}} &
  \multicolumn{1}{l}{1.2 million} &
  \multicolumn{1}{l}{varies} &
  \multicolumn{1}{l}{Scene recognition} &
  \multicolumn{1}{l}{Low} \\ \cmidrule(l){2-6} 
\multicolumn{1}{l}{} &
  \multicolumn{1}{l}{MNIST~\cite{lecun1998gradient}} &
  \multicolumn{1}{l}{70,000} &
  \multicolumn{1}{l}{784} &
  \multicolumn{1}{l}{Object recognition} &
  \multicolumn{1}{l}{High} \\ \midrule
\multicolumn{1}{l}{\multirow{4}{*}{Text}} &
  \multicolumn{1}{l}{IMDB~\cite{maas-EtAl:2011:ACL-HLT2011}} &
  \multicolumn{1}{l}{50,000} &
  \multicolumn{1}{l}{varies} &
  \multicolumn{1}{l}{Sentiment analysis} &
  \multicolumn{1}{l}{Medium} \\ \cmidrule(l){2-6} 
\multicolumn{1}{l}{} &
  \multicolumn{1}{l}{Newsgroup~\cite{joachims1996probabilistic}} &
  \multicolumn{1}{l}{19,188} &
  \multicolumn{1}{l}{varies} &
  \multicolumn{1}{l}{Text classification} &
  \multicolumn{1}{l}{Low} \\ \cmidrule(l){2-6} 
\multicolumn{1}{l}{} &
  \multicolumn{1}{l}{Reuters~\cite{apte1994automated}} &
  \multicolumn{1}{l}{10,788} &
  \multicolumn{1}{l}{varies} &
  \multicolumn{1}{l}{Text classification} &
  \multicolumn{1}{l}{Low} \\ \cmidrule(l){2-6} 
\multicolumn{1}{l}{} &
  \multicolumn{1}{l}{SQuAD~\cite{rajpurkar2016squad}} &
  \multicolumn{1}{l}{100,000} &
  \multicolumn{1}{l}{Varies} &
  \multicolumn{1}{l}{Question answering} &
  \multicolumn{1}{l}{Low} \\ \midrule
\multicolumn{1}{l}{\multirow{3}{*}{Tabular}} &
  \multicolumn{1}{l}{Adult~\cite{kohavi1996scaling}} &
  \multicolumn{1}{l}{48,842} &
  \multicolumn{1}{l}{14} &
  \multicolumn{1}{l}{Income prediction} &
  \multicolumn{1}{l}{Low} \\ \cmidrule(l){2-6} 
\multicolumn{1}{l}{} &
  \multicolumn{1}{l}{Breast Cancer~\cite{michalski1986multi}} &
  \multicolumn{1}{l}{286} &
  \multicolumn{1}{l}{9} &
  \multicolumn{1}{l}{Cancer diagnosis} &
  \multicolumn{1}{l}{Low} \\ \cmidrule(l){2-6} 
\multicolumn{1}{l}{} &
  \multicolumn{1}{l}{Diabetes~\cite{smith1988using}} &
  \multicolumn{1}{l}{768} &
  \multicolumn{1}{l}{8} &
  \multicolumn{1}{l}{Diabetes diagnosis} &
  \multicolumn{1}{l}{Low} \\ \midrule
\multicolumn{1}{l}{\multirow{2}{*}{Time series}} &
  \multicolumn{1}{l}{Epileptic Seizure~\cite{andrzejak2001indications}} &
  \multicolumn{1}{l}{11,500} &
  \multicolumn{1}{l}{178} &
  \multicolumn{1}{l}{Seizure prediction} &
  \multicolumn{1}{l}{Low} \\ \cmidrule(l){2-6} 
\multicolumn{1}{l}{} &
  \multicolumn{1}{l}{Activity Recognition~\cite{anguita2013public}} &
  \multicolumn{1}{l}{10,299} &
  \multicolumn{1}{l}{561} &
  \multicolumn{1}{l}{Activity Classification} &
  \multicolumn{1}{l}{Low} \\ \midrule
\multicolumn{1}{l}{\multirow{3}{*}{Graph}} &
  \multicolumn{1}{l}{OGB~\cite{hu2020open}} &
  \multicolumn{1}{l}{1.2 million} &
  \multicolumn{1}{l}{varies} &
  \multicolumn{1}{l}{Graph classification} &
  \multicolumn{1}{l}{Low} \\ \cmidrule(l){2-6} 
\multicolumn{1}{l}{} &
  \multicolumn{1}{l}{Cora~\cite{sen2008collective}} &
  \multicolumn{1}{l}{2,708} &
  \multicolumn{1}{l}{1,433} &
  \multicolumn{1}{l}{Graph classification} &
  \multicolumn{1}{l}{Low} \\ \cmidrule(l){2-6} 
\multicolumn{1}{l}{} &
  \multicolumn{1}{l}{Yelp Dataset~\cite{zhang2015character}} &
  \multicolumn{1}{l}{8,282,442} &
  \multicolumn{1}{l}{Varies} &
  \multicolumn{1}{l}{Recommendation} &
  \multicolumn{1}{l}{Low} \\ \midrule
\multicolumn{1}{l}{\multirow{5}{*}{Computer Vision}} &
  \multicolumn{1}{l}{Fashion-MNIST~\cite{xiao2017fashion}} &
  \multicolumn{1}{l}{70,000} &
  \multicolumn{1}{l}{784} &
  \multicolumn{1}{l}{Image classification} &
  \multicolumn{1}{l}{Medium} \\ \cmidrule(l){2-6} 
\multicolumn{1}{l}{} &
  \multicolumn{1}{l}{Caltech-101~\cite{fei2004learning}} &
  \multicolumn{1}{l}{9,146} &
  \multicolumn{1}{l}{Varies} &
  \multicolumn{1}{l}{Object recognition} &
  \multicolumn{1}{l}{Low} \\ \cmidrule(l){2-6} 
\multicolumn{1}{l}{} &
  \multicolumn{1}{l}{COCO~\cite{lin2014microsoft}} &
  \multicolumn{1}{l}{330,000} &
  \multicolumn{1}{l}{Varies} &
  \multicolumn{1}{l}{Object detection} &
  \multicolumn{1}{l}{Medium} \\ \cmidrule(l){2-6} 
\multicolumn{1}{l}{} &
  \multicolumn{1}{l}{YouTube Faces~\cite{wolf2011face}} &
  \multicolumn{1}{l}{3,425} &
  \multicolumn{1}{l}{2,622} &
  \multicolumn{1}{l}{Face recognition} &
  \multicolumn{1}{l}{Medium} \\ \cmidrule(l){2-6} 
\multicolumn{1}{l}{} &
  \multicolumn{1}{l}{EuroSAT~\cite{helber2019eurosat}} &
  \multicolumn{1}{l}{27,000} &
  \multicolumn{1}{l}{13} &
  \multicolumn{1}{l}{Land use classification} &
  \multicolumn{1}{l}{Low} \\ \midrule
\multicolumn{1}{l}{Transaction} &
  \multicolumn{1}{l}{Purchase~\cite{sakar2019real}} &
  \multicolumn{1}{l}{39,624} &
  \multicolumn{1}{l}{8} &
  \multicolumn{1}{l}{Purchase prediction} &
  \multicolumn{1}{l}{Medium} \\ \midrule
\multicolumn{1}{l}{Sequence} &
  \multicolumn{1}{l}{\begin{tabular}[c]{@{}l@{}}Human Activity\\ Recognition~\cite{anguita2012human}\end{tabular}} &
  \multicolumn{1}{l}{10,299} &
  \multicolumn{1}{l}{561} &
  \multicolumn{1}{l}{Activity recognition} &
  \multicolumn{1}{l}{Low} \\ \midrule
\multicolumn{1}{l}{Recommendation} &
  \multicolumn{1}{l}{MovieLens~\cite{10.1145/2827872}} &
  \multicolumn{1}{l}{100,000} &
  \multicolumn{1}{l}{varies} &
  \multicolumn{1}{l}{Movie recommendation} &
  \multicolumn{1}{l}{High} \\ \bottomrule
\end{tabular}}
\footnotesize\textsuperscript{*} Refer to Supplementary material.
\end{table}

\section{Datasets and Metrics for Evaluation}\label{evaluation}

The effectiveness of MU techniques is assessed using different metrics, including accuracy on forget and retain sets, relearn time, MI attacks, activation distance, AIN, and layer-wise distance. This section offers a brief overview of these metrics. It addresses the second research question ``How can the impact of unlearning on model performance be measured and evaluated?'' as shown in Tab.~\ref{tab:evaluation}, along with a summary of publicly available datasets including their types, number of instances, attributes and tasks, and their popularity based on number of references they appear in. These are presented in Tab.~\ref{tab:datasets}.

The metrics included in Tab.~\ref{tab:evaluation} are commonly used in MU research to evaluate the effectiveness of different MU techniques. By comparing the values of these metrics before and after applying the MU techniques, researchers can assess the degree to which sensitive data has been removed or mitigated.

\begin{table*}[]
\centering
\caption{Machine Unlearning - Evaluation Metrics}
\label{tab:evaluation}
\resizebox{\textwidth}{!}{%
\begin{tabular}{@{}llll@{}}
\toprule
\textbf{Metric} &
  \textbf{Description} &
  \multicolumn{1}{l}{\textbf{Equation}} &
  \textbf{Reference} \\ \midrule
\multicolumn{1}{l}{Accuracy} &
  \multicolumn{1}{l}{The proportion of correctly classified instances in the test set} &
  \multicolumn{1}{l}{$Accuracy = \frac{TP + TN}{TP + TN + FP + FN}$} &
  \multicolumn{1}{l}{~\cite{9157084}} \\ \midrule
\multicolumn{1}{l}{Anamnesis Index} &
  \multicolumn{1}{l}{Measures the extent to which unlearned data has been forgotten} &
  \multicolumn{1}{l}{$AI(T) = (Acc(M, T) - Acc(M-T, T)) / Acc(Naive, T)$} &
  \multicolumn{1}{l}{~\cite{parisi2019continual}} \\ \midrule
\multicolumn{1}{l}{Activation Distance} &
  \multicolumn{1}{l}{\begin{tabular}[c]{@{}l@{}}Measures the difference between the activations of two models\\ on a given input\end{tabular}} &
  \multicolumn{1}{l}{$AD(A(x), B(x)) = ||A(x) - B(x)||_{2}$} &
  \multicolumn{1}{l}{~\cite{9157084}} \\ \midrule
\multicolumn{1}{l}{Layer-wise Distance} &
  \multicolumn{1}{l}{\begin{tabular}[c]{@{}l@{}}Measures the difference between the weight \\ matrices of two models\end{tabular}} &
  \multicolumn{1}{l}{$LWD_l = ||W_l - W'_l||_F$} &
  \multicolumn{1}{l}{~\cite{https://doi.org/10.48550/arxiv.2111.08947}} \\ \midrule
\multicolumn{1}{l}{\begin{tabular}[c]{@{}l@{}}Membership \\ Inference Attacks\end{tabular}} &
  \multicolumn{1}{l}{\begin{tabular}[c]{@{}l@{}}Measures the degree to which an attacker can infer whether a \\ particular instance was included in the training data\end{tabular}} &
  \multicolumn{1}{l}{N/A} &
  \multicolumn{1}{l}{~\cite{Golatkar2021}} \\ \midrule
\multicolumn{1}{l}{\begin{tabular}[c]{@{}l@{}}Zero Retrain \\ Forgetting Metric\end{tabular}} &
  \multicolumn{1}{l}{\begin{tabular}[c]{@{}l@{}}Measures the change in accuracy on the unlearned data after \\ retraining the model without the data to be unlearned\end{tabular}} &
  \multicolumn{1}{l}{$ZRFM = \frac{1}{L}\sum_{l=1}^{L}\left(1 - \frac{|W_{l}^{T}W_{l} - W_{l}^{T}{old}W{l,old}|{F}}{|W{l}^{T}W_{l,old}|_{F}}\right)$} &
  \multicolumn{1}{l}{~\cite{10.1145/3488932.3517406}} \\ \midrule
\multicolumn{1}{l}{Reconstruction Error}&
  \begin{tabular}[c]{@{}l@{}}Measures the difference between the original input and\\  the reconstructed input\end{tabular} &
  $RE = ||x - f(g(x))||2$ &
  ~\cite{tan2023unfolded} \\ \bottomrule
\end{tabular}}
\end{table*}

\section{Challenges and Potential Solutions}\label{challenges}
\begin{figure}
    \centering
    \includegraphics[width=\columnwidth]{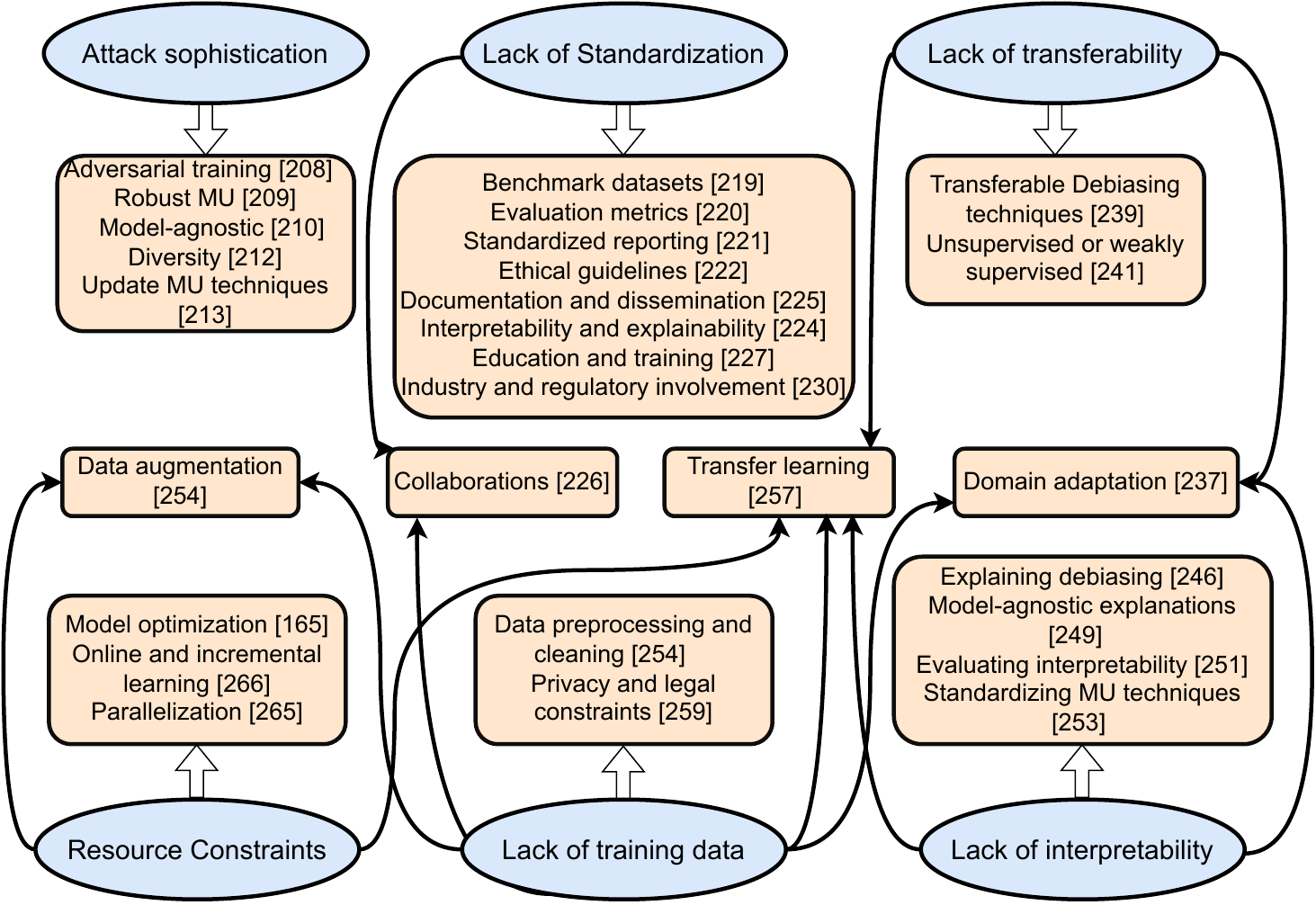}
    \caption{Machine Unlearning Challenges and Potential Solutions - A Roadmap}
    \label{fig:challenges}
\end{figure}

This section discusses the prevalent challenges in the field of MU and the 
potential solutions to these challenges in detail, which aims to address the third research question ``\textit{What are the challenges in MU, and how can these
challenges be addressed?}''. A roadmap of the challenges and their corresponding solutions can be found in Fig.~\ref{fig:challenges}.

\subsection{Attack sophistication} As MU techniques become more sophisticated, so too do the potential attacks designed to compromise them. Attackers may use a combination of different techniques, such as data poisoning, model inversion or adversarial attacks, to try to undermine the privacy and security of ML models~\cite{lin2021ml}. The challenge of attack sophistication in MU refers to the potential for adversaries to develop sophisticated techniques to bypass or defeat MU defenses. As MU techniques are developed to mitigate the impact of biased or adversarial data on models, adversaries may adapt and evolve their attack strategies to counter these defenses. This brings us to the following set of existing challenges:

\begin{itemize}
    \item \textit{Increasing sophistication of attacks:} Adversaries can develop sophisticated techniques to generate biased or adversarial data that can evade detection by MU techniques. For example, they can craft adversarial examples that are specifically designed to bypass the debiasing or unlearning process, making it challenging for the defenses to effectively identify and mitigate the biases~\cite{carlini2021extracting}.
    \item \textit{Adaptive adversaries:} Adversaries can adapt their attack strategies based on the observed behavior of MU defences~\cite{gupta2021adaptive, sommer2020towards, warnecke2021machine}. If they notice that certain debiasing techniques are applied, they may modify their attacks to target the specific vulnerabilities or limitations of those techniques. This can lead to a cat-and-mouse game, where adversaries continually adapt their attacks, making it difficult for MU defenses to keep up.
    \item \textit{Stealthy attacks:} Adversaries can also develop stealthy attacks that are designed to remain undetected by MU defenses~\cite{guo2022backdoor}. These attacks may not necessarily generate overtly biased or adversarial data, but instead subtly manipulate the data in a way that evades detection by the defences~\cite{zheng2023one4all,guo2023scale}. This can make it challenging for MU techniques to effectively identify and address the biases or adversarial influences in the data~\cite{yu2022don}.
    \item \textit{Lack of robustness:} MU techniques may have limitations in terms of their robustness against sophisticated attacks~\cite{schelter2021hedgecut,zhang2022poison}. For example, certain debiasing or unlearning methods may be effective against simple attacks, but may not be robust enough to withstand more sophisticated attacks~\cite{madva2017biased}. This can result in the effectiveness of MU defenses being compromised by increasingly sophisticated attacks.
\end{itemize}

Addressing the challenge of attack sophistication requires the development of robust and adaptive defenses that can effectively detect and mitigate sophisticated attacks. This may involve continuously updating and improving MU techniques with knowledge of evolving attack strategies, as well as incorporating techniques from areas such as adversarial ML model and robust ML model. 
\begin{itemize}
\item \textit{Adversarial training} is a technique that involves training the ML model on adversarial examples. This can improve the model's robustness against adversarial attacks~\cite{madry2017towards}. Adversarial training can also be used in conjunction with MU techniques to make the model more resilient against sophisticated attacks.
\item \textit{Robust MU techniques} address the need to develop MU techniques that are robust against sophisticated attacks. For instance, MU techniques can be combined with adversarial training to improve their effectiveness against adaptive adversaries~\cite{wu2020adversarial}. This can make it more challenging for adversaries to bypass the debiasing or unlearning process.
\item \textit{Model-agnostic techniques} can be used to detect and mitigate biases in ML models, without relying on the specifics of the model architecture~\cite{zemel2013learning,pleiss2017fairness}. These techniques can be more robust against sophisticated attacks because they do not rely on assumptions about the model architecture.
\item \textit{Diversity in the training data} can make the ML model more robust against attacks~\cite{noack2021empirical}. For instance, including data from a wider range of sources and demographics can make it more challenging for adversaries to generate biased or adversarial examples that can bypass the debiasing or unlearning process.
\item \textit{Regular re-evaluation and updating of MU techniques} 
can help to stay ahead of attackers and keep the defenses effective against the adapting attack strategies of adversaries~\cite{wachter2017transparent}.
\end{itemize}


\subsection{Lack of standardization} There is currently no standardization in the application of MU techniques, making it difficult to compare different approaches and evaluate their effectiveness. The lack of standardization results in a current gap of commonly accepted standards, guidelines, or best practices for unlearning techniques.
And because MU is relatively new and still evolving, there is presently a lack of standardized approaches, evaluation metrics, or benchmarks for evaluating the effectiveness, fairness, and robustness of unlearning methods~\cite{tahiliani2021machine}. This lack of standardization can pose challenges across several areas:

\begin{itemize}
    \item \textit{Methodology:} There is no universally accepted methodology for unlearning, and different researchers or practitioners may use different approaches, techniques or algorithms for unlearning biases or unwanted behaviors in ML models. This lack of standardization makes it difficult to compare and reproduce results across studies or applications, and may lead to inconsistent or unreliable outcomes~\cite{tahiliani2021machine}.
    \item \textit{Evaluation:} The evaluation of unlearning techniques is challenging due to the absence of standardized evaluation metrics or benchmarks. It is often unclear how to objectively measure the effectiveness, fairness or robustness of unlearning methods, and different studies may use different evaluation metrics or criteria, making it difficult to compare and assess the performance of different unlearning techniques~\cite{liu2022right}.
    \item \textit{Data and model compatibility:} Unlearning techniques may need to be compatible with different types of data, ML models, or applications. However, there is currently no standardized framework for ensuring data and model compatibility~\cite{achille2023ai}, meaning unlearning methods must be adapted or customized for different use cases or applications, which can be time-consuming and resource intensive.
    \item \textit{Ethical considerations:} Unlearning techniques may raise ethical concerns, such as issues related to fairness, accountability, transparency or unintended consequences~\cite{pradhan2022interpretable}. However, there is currently no standardized framework or guidelines for addressing ethical considerations in MU, and different practitioners or researchers may have different perspectives or approaches toward ethical considerations. This may lead to potential inconsistencies or biases in unlearning outcomes~\cite{warnecke2021machine}.
    \item \textit{Interpretability and explainability:} Unlearning techniques may result in modified or updated models, and it is important to understand and explain the changes made to the original models~\cite{thudi2022necessity}. However, there is currently no standardized approach for interpreting or explaining the changes made by unlearning methods, which can make it challenging to understand, trust or communicate the outcomes of unlearning techniques to stakeholders.
\end{itemize}
The lack of standardization challenge requires efforts from the researchers and practitioners to establish common frameworks, guidelines, and best practices. Some potential approaches to promote standardization in the MU community include:
\begin{itemize}
    \item \textit{Community-wide collaborations:} Encouraging collaboration and communication among researchers, practitioners and stakeholders to establish shared standards, guidelines and best practices for MU~\cite{vu2022mindful}. This can involve organizing MU-dedicated workshops, conferences or forums to foster discussions and promote the exchange of ideas and experiences.
    \item \textit{Benchmark datasets and evaluation metrics:} Developing benchmark datasets and evaluation metrics for unlearning techniques will provide a standardized means to objectively evaluate the performance, fairness, robustness, and other aspects of unlearning methods~\cite{moon2023feature}. These benchmarks can help compare and assess the effectiveness of different unlearning techniques in a consistent and reproducible manner.
    \item \textit{Standardized reporting:} There is a need to encourage standardized reporting of unlearning studies, including clear descriptions of methodologies, algorithms, data, and the evaluation metrics used~\cite{Park2020}. This can ensure transparency, reproducibility and comparability of unlearning research findings, allowing for a better understanding and assessment of the effectiveness of different unlearning techniques.
    \item \textit{Ethical guidelines:} There is a need to develop ethical guidelines for MU, addressing issues such as fairness, accountability, transparency, and unintended consequences~\cite{Sand2021}. Ethical guidelines can provide a framework for practitioners to consider and address ethical concerns associated with unlearning techniques, and promote responsible and ethical use of unlearning methods~\cite{Tth2022}.
    \item \textit{Interpretability and explainability:} Developing standardized approaches for interpreting and explaining the outcomes of unlearning techniques, such as methods for model introspection, visualization or explanation~\cite{sajid2023methodology}, can enhance transparency, trust, and accountability. It will also further facilitate effective communication of unlearning results to stakeholders.
    \item \textit{Documentation and dissemination:} The domain must encourage its researchers and practitioners to document and share their unlearning methodologies, techniques and results through open repositories, code libraries, or other means of dissemination~\cite{takeuchi2021learning}. This can promote the exchange of knowledge, facilitate replication and foster the development of shared practices and standards in the field of MU.
    \item \textit{Collaborative efforts with other fields:} Collaborating with related fields, such as ML, fairness, ethics, or interpretability researchers, will leverage existing standards, methodologies, or frameworks~\cite{hanif2021survey}. Drawing upon established practices from related fields can help develop standardized approaches for MU and promote cross-disciplinary collaboration.
    \item \textit{Education and training:} Incorporating MU concepts, methodologies, and best practices into educational programs, training workshops, or industry certifications, particularly through education and training on standardized approaches, can help practitioners develop a common understanding and adoption of best practices in the MU field~\cite{Bhler2022, Pvloaia2023}.
    \item \textit{Industry and regulatory involvement:} Industry and regulatory bodies should be actively involved in the development of standards, guidelines, and best practices for MU~\cite{Manning2023}. Industry participation will ensure practical relevance and adoption of standards, while regulatory involvement can promote responsible and ethical use of MU techniques~\cite{Kumar2021}.
\end{itemize}


\subsection{Lack of transferability} 
While MU is an important technique for addressing bias and improving fairness in ML, one challenge associated with MU is the lack of transferability~\cite{fletcher2021addressing}. Some MU techniques may be effective for specific types of models or datasets, but may not be easily transferable to other models or datasets~\cite{ross2018improving}. This can limit their applicability in real-world scenarios. This challenge arises due to several reasons:

\begin{itemize}
    \item \textit{Model-specific biases:} Different ML models may exhibit different types or degrees of bias~\cite{brophy2020exit}, and unlearning techniques that are effective for one model may not be directly applicable to another model. For example, a debiasing method designed for a DNN may not be directly transferable to a decision tree or a support vector machine~\cite{buhrmester2021analysis}.
    \item \textit{Dataset-specific biases:} Unlearning techniques may be tailored to address biases in specific datasets, and their effectiveness may depend on the characteristics of the data used for training~\cite{ashraf2018learning, sarkar2021trapdoor}. When applied to a different dataset, the unlearning techniques may not be as effective, as the biases present in the new dataset may be different or require different mitigation strategies.
    \item \textit{Domain-specific biases:} Unlearning techniques may be developed and trained on data from a specific domain, such as healthcare, finance, or social media, and may not be directly transferable to other domains~\cite{dinsdale2022fedharmony, hwang2020variational}. Different domains have unique characteristics, data distributions, and biases, which may require domain-specific adaptations of MU techniques.
    \item \textit{Lack of labeled data:} Unlearning techniques may require labeled data for training, validation,\textit{} or fine-tuning, meaning that obtaining labeled data for new models, datasets, or domains may be challenging or costly~\cite{ma2022learn}. The lack of labeled data can hinder the transferability of unlearning techniques, as the models may not have access to the same level of labeled data for training as the original models.
\end{itemize}
MU requires careful consideration of the specific context and characteristics of the models, datasets or domains to address the lack of transferability. Some potential approaches to mitigate this challenge include:
\begin{itemize}
    \item \textit{Domain adaptation} involves incorporating techniques such as domain adaptation algorithms or domain generalization methods~\cite{yin2020speaker}, to adapt unlearning techniques to new domains. These techniques involve leveraging knowledge from a source domain to improve the performance of the unlearning techniques in a target domain, even when labeled data is limited to the target domain~\cite{corral20212}.
    \item \textit{Transfer learning} involves employing transfer learning techniques, such as transfer learning algorithms or pre-trained models, to leverage knowledge learned from one model or dataset to another. Transfer learning can help transfer the learned debiasing or unlearning knowledge from one model to another, even when the models have different architectures or characteristics.
    \item \textit{Transferable debiasing techniques} is developing debiasing techniques that are designed to be transferable across different models, datasets, or domains. These techniques may incorporate generalization principles, domain-independent features, or model-agnostic approaches that enable their application to diverse settings~\cite{wang2023large,zhao2022using}.
    \item \textit{Unsupervised or weakly supervised approaches} involve exploring unsupervised or weakly supervised approaches that do not require labeled data for training. Unsupervised or weakly supervised unlearning techniques may leverage unsupervised learning, self-supervised learning, or weakly supervised learning methods to overcome the lack of labeled data and enhance transferability~\cite{zhang2022survey,cheng-etal-2021-posterior}.
    \item\textit{ Benchmark datasets} to span different models, datasets, or domains, and evaluating unlearning techniques on these benchmark datasets~\cite{kalinowska2023embodied,goldblum2022dataset}. Benchmark datasets can facilitate standardized evaluations of unlearning techniques across diverse settings and provide insights into their transferability and generalizability.
\end{itemize}


\subsection{Lack of interpretability} 
MU techniques may be difficult to interpret, particularly those involving complex DNNs or generative models. Difficulties include understanding how the technique is working and identifying potential weaknesses or vulnerabilities. Interpretability refers to the ability to understand and explain how an ML model makes predictions or decisions~\cite{brophy2021machine}. It is crucial for building trust, accountability, and transparency in ML systems, as it allows stakeholders to understand and verify the reasoning behind model outputs. However, MU techniques can sometimes result in models that are less interpretable for several reasons, such as the complexity of unlearning techniques, loss of transparency during debiasing, lack of standardization, and trade-offs between accuracy and interpretability. 

Addressing this challenge requires careful consideration of the specific de-biasing or unlearning techniques used and their impact on model interpretability. Some potential approaches to enhance interpretability include:
\begin{itemize}
    \item \textit{Explaining the debiasing process} by providing clear explanations on the data modifications or feature changes during unlearning can help stakeholders understand the reasoning behind the model's outputs~\cite{bevan2022skin,tarun2022deep,he2019unlearn}.
    \item \textit{Simplifying unlearning techniques} can develop simpler and more interpretable MU techniques that are easier to understand and explain to stakeholders~\cite{chien2022certified}.
    \item \textit{Providing model-agnostic explanations} such as feature importance~\cite{lu2017feature} or partial dependence plots, can help provide insights into how different features or data points influence the model's predictions, regardless of the specific MU technique used ~\cite{jia2022zest}.
    \item \textit{Evaluating interpretability} alongside accuracy by incorporating interpretability as an explicit evaluation criterion during the development and assessment of MU techniques, to help ensure that interpretability is not compromised in the pursuit of accuracy~\cite{mohseni2021quantitative,mohseni2018human}.
    \item \textit{Standardizing MU techniques} such as standardized approaches and guidelines for implementing and evaluating MU methods, can help promote consistency, reproducibility, and interpretability in the field~\cite{kaur2022trustworthy}.
\end{itemize}

\subsection{Lack of training data} 
In some cases, it may be difficult to obtain sufficient training data for MU techniques, particularly in scenarios where the original training data may be scarce or difficult to obtain~\cite{saunders2022domain}. The availability and quality of training data are critical factors that impact the effectiveness of MU techniques, which face significant challenges otherwise. Some challenges associated with the lack of training data in MU include scarce biased or adversarial data, data imbalance, privacy and legal constraints, data quality issues, and domain shifts~\cite{osuala2021review}. Addressing the lack of training data challenges in MU requires careful consideration of the specific techniques used and their data requirements. Some potential approaches to mitigate this challenge include:
\begin{itemize}
    \item \textit{Data augmentation} such as data synthesis or generation methods to augment the available training data can help address data scarcity and imbalance challenges~\cite{Strelcenia2023}. This can involve generating synthetic data to supplement the biased or adversarial data, and oversampling minority or underrepresented groups to address data imbalance~\cite{Rana2022}.
    \item \textit{Data pre-processing and cleaning} ensures the quality and accuracy of training data, which can help address data quality issues~\cite{Fu2020}. This involves removing noise, inaccuracies, or inconsistencies in the training data to improve the reliability and effectiveness of MU techniques~\cite{NEURIPS2022_5771d9f2}.
    \item \textit{Transfer learning or domain adaptation} techniques can help address the challenge of domain shift~\cite{Bu2020}. These techniques involve leveraging knowledge from a source domain with abundant data to improve the performance of the model in a target domain with limited data~\cite{Fu2020}.
    \item \textit{Collaborative efforts and data sharing} especially among researchers, organizations, and stakeholders, can help overcome data availability challenges by pooling resources, sharing data, and collectively addressing data limitations~\cite{Kar2020}. This can lead to the creation of shared datasets, benchmark datasets, and other data resources for use in MU research and development.
    \item \textit{Compliance with privacy and legal constraints} is crucial~\cite{Aljeraisy2021}. Researchers and practitioners should carefully follow ethical guidelines, data protection regulations, and legal requirements to ensure that MU techniques are applied in a responsible and compliant manner. This may involve obtaining necessary permissions, anonymizing data, or adhering to data-sharing agreements or data use policies.
\end{itemize}

\subsection{Resource constraints} Resource constraints are a challenge in MU, referring to limitations in terms of computational resources, time, or data availability, which can impact the effectiveness and efficiency of unlearning techniques~\cite{maddikunta2022incentive}. MU methods can be computationally intensive, time-consuming, or data-dependent, and resource constraints pose significant challenges in their practical implementation~\cite{sekhari2021remember}. Some common types of resource constraints in MU include:
\begin{itemize}
    \item \textit{Computational resources} such as processing power, memory, or storage, may be intensive when required to effectively train or apply debiasing models~\cite{Tuladhar2022}. Several deep learning-based unlearning methods may require large-scale neural networks or extensive computational resources for training, which may not be readily available in resource-constrained environments or in applications with low-power devices~\cite{Rathi2023}.
    \item \textit{Time constraints} for when unlearning techniques may require significant time to train or apply, especially when dealing with large datasets or complex models~\cite{Graves2021}. Time constraints can be particularly challenging in real-time or online learning scenarios, where timely debiasing or unlearning is critical~\cite{Mondal2022}. For example, in time-sensitive applications such as fraud detection or online advertising, the time required for unlearning may impact overall system performance or responsiveness.
    \item \textit{Data availability} particularly for labeled data can be costly in scenarios where unlearning techniques require labeled data for training, validation, or fine-tuning~\cite{NIPS2014_375c7134}. Data availability constraints may arise when dealing with sensitive or proprietary data, or when data collection is time-consuming or expensive~\cite{Hsu2020}. Lack of data availability can impact the effectiveness and generalizability of unlearning techniques, as they may not have access to sufficient data to effectively mitigate biases.
\end{itemize} 

Addressing resource constraints requires careful consideration of the available resources and the specific requirements of the unlearning techniques. Some potential approaches to mitigate resource constraints in MU include:
\begin{itemize}
    \item \textit{Model optimization} for the computational resources used by unlearning techniques by using techniques such as model compression, model quantization or model approximation. These techniques can reduce the computational overhead of unlearning methods and enable their deployment in resource-constrained environments or on low-power devices~\cite{Aldaghri2021,chen2021machine}.
    \item \textit{Parallelization} to leverage parallel computing techniques, such as distributed computing, multi-core processors or GPUs, can accelerate the training or application of unlearning techniques. Parallelization can help overcome computational resource constraints and expedite the debiasing or unlearning process~\cite{yuan2023federated}~\cite{li2023subspace}.
    \item \textit{Online and incremental learning approaches} that update the debiasing or unlearning models in an online or incremental manner, without requiring complete retraining~\cite{cauwenberghs2000incremental}. Online and incremental learning can be more time-efficient and resource-friendly compared to batch training methods, as they update the models incrementally based on new data rather than retrain the entire model from scratch~\cite{romero2007incremental,tsai2014incremental}.
    \item \textit{Transfer learning techniques} such as fine-tuning or domain adaptation, to leverage pre-trained models or knowledge from related tasks or domains~\cite{xu2020transfer}. Transfer learning can help overcome data availability constraints by leveraging knowledge learned from other sources, reducing the reliance on labeled data for training unlearning models.
    \item \textit{Data augmentation techniques} such as data synthesis, data generation, or data simulation, to augment the available data and create synthetic data points for training unlearning models. Data augmentation can help overcome the challenge by generating additional training data, which can improve the effectiveness and generalizability of unlearning techniques~\cite{deepanjali2021efficient,sommer2020towards,di2022hidden}.
\end{itemize}

\section{Future Directions}\label{future}

\subsection{Machine Unlearning in NLP}
One of the primary goals of ML models in NLP is to accurately predict or classify text data based on patterns and relationships learned from a training dataset~\cite{shaik2022review}. However, if the training data is incomplete, biased, or outdated, the model may learn incorrect or irrelevant patterns, leading to poor performance when the model is applied to new data. Unlearning is a way to correct or remove these incorrect patterns, which can improve the model's performance on new data~\cite{mehta2022deep}. For example, if a model is trained to classify news articles as either political or non-political based on the presence of certain keywords but the model has learned to rely too heavily on a specific keyword that is no longer relevant or has changed in meaning, the model may need to unlearn its reliance on that keyword to make more accurate predictions. Removing irrelevant or outdated patterns from the model can improve its ability to generalize to new data and make accurate predictions~\cite{maharana2022review}. In some cases, unlearning may involve retraining the model on a new dataset that includes examples that are deliberately chosen to contradict or challenge the original model learning~\cite{stacey2022supervising}. This can help the model adapt to new information and improve its accuracy and generalizability.



\subsection{Machine Unlearning in Computer Vision}
The future of MU in Computer Vision (CV) is also likely to be influenced by continued advancements in ML techniques, and an increasing emphasis on interpretability and transparency. One area where MU may play a significant role in the future of CV is in the development of more robust and adaptable models~\cite{sarker2022multi}. As the volume and complexity of visual data continues to grow, ML models will need to be able to adapt to new types of data and learn from changing user needs. Another promising research area is incremental learning, which involves updating a model over time as new data becomes available. This approach could be particularly useful in scenarios where the model needs to adapt to changing conditions, such as in autonomous vehicles or surveillance systems~\cite{lobo2020spiking}. An important development in CV is the move towards more explainable AI, which would allow users to understand how a model arrived at a particular decision and potentially correct errors or biases in the process~\cite{sajid2023methodology}. In some cases, it may be necessary to involve human experts in the unlearning process, either to identify errors in the model or to provide feedback on the effectiveness of the unlearning approach~\cite{du2019lifelong}. As MU becomes more widespread, it will be important to ensure that sensitive data is protected and that unlearning processes do not compromise security or privacy~\cite{fan2023machine}. This may require the development of new techniques and approaches to secure MU processes.

\subsection{Machine Unlearning in Recommender Systems}
MU in Recommender Systems (RS) is likely to be characterized by continued advancements in ML techniques, as well as an increasing focus on interpretability, fairness, and privacy~\cite{kaissis2020secure}. One area where MU may play an important role in the future of RS is the development of more adaptive and personalized models~\cite{chen2022recommendation}. As the volume and complexity of user data grows, MU models will need to be able to adapt to changing user preferences and needs. Unlearning can help these models stay up-to-date and make accurate recommendations by allowing them to adjust or remove irrelevant or misleading features~\cite{ramscar2021discriminative}. Another area where MU may be important for RS is in the development of more transparent and fair models~\cite{yin2023trustworthy}. MU can help to improve the interpretability and explainability of RS~\cite{yao2022counterfactually}. Unlearning can help to make these models more transparent by allowing them to remove irrelevant or misleading features and focus on the most important factors for making accurate recommendations~\cite{asatiani2021sociotechnical}. Finally, MU can help to protect user privacy in RS. By removing or obfuscating sensitive or identifying information from the data used to train these models, MU can ensure that a user's personal information is not used inappropriately or disclosed without their consent~\cite{schnitzler2023analyzing}.

\section{Conclusion}\label{conclusion}

MU is a relatively new and rapidly evolving field that has gained increasing attention in recent years. While the process of training ML models to recognize patterns and make predictions has become increasingly efficient and accurate, the need to remove or modify these predictions has become equally important. Unlearning, as the name implies, refers to the process of removing previously learned information from a model, and it has important applications in areas such as privacy, security, and fairness. As our literature survey has shown, there are a variety of approaches and techniques being developed for unlearning data, ranging from regularization methods to model inversion techniques. However, there are still challenges that need to be addressed in this area, such as scalability to larger datasets, the ability to unlearn specific subsets of data, and the impact of unlearning on model performance. However, despite these challenges, the benefits of MU are significant, and we expect to see continued progress in this field in the coming years as researchers develop more effective and efficient methods for unlearning data from ML models. Researchers and practitioners must continue to explore and refine unlearning techniques to ensure that ML models can to adapt to changing circumstances and maintain the trust of their users. With the increasing importance of AI in various domains, unlearning will play a crucial role in making AI more trustworthy and transparent.

\bibliographystyle{ieeetr}
\bibliography{bare_jrnl_new_sample4}
\vfill
\begin{IEEEbiography}[{\includegraphics[width=1in,height=1.25in,clip,keepaspectratio]{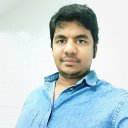}}]{Thanveer Shaik} received the master’s degree in applied data science from the University of Southern Queensland, Australia, where he is currently pursuing the Ph.D. degree. His research interests include cognitive computing, biometrics, and NLP with expertise in artificial intelligence (AI), machine learning, and predictive analysis. He published papers in prestige Journals (e.g., KBS, WIREs DMKD, Sensors, IEEE Access, and RPTEL, etc.).
\end{IEEEbiography}
\vskip -3\baselineskip plus -1fil
\begin{IEEEbiography}[{\includegraphics[width=1in,height=1.25in,clip,keepaspectratio]{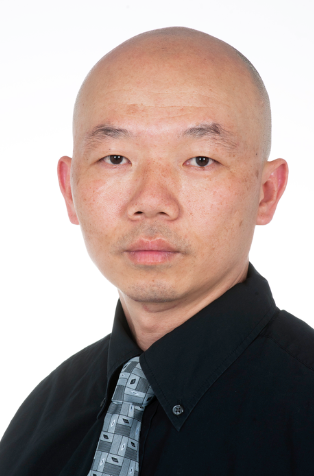}}]{Xiaohui Tao (Senior Member, IEEE)} is Full Professor with University of Southern Queensland, Australia. His research interests cover data analytics, natural language processing, machine learning, knowledge engineering, and health informatics. His research outcomes have been published on many top-tier journals (e.g., TKDE, INFFUS, IPM) and conferences (e.g., AAAI, IJCAI, ICDE, CIKM). Tao is Vice Chair of IEEE Technical Committee of Intelligent Informatics.
\end{IEEEbiography}
\vskip -3\baselineskip plus -1fil
\begin{IEEEbiography}[{\includegraphics[width=1in,height=1.25in, clip,keepaspectratio]{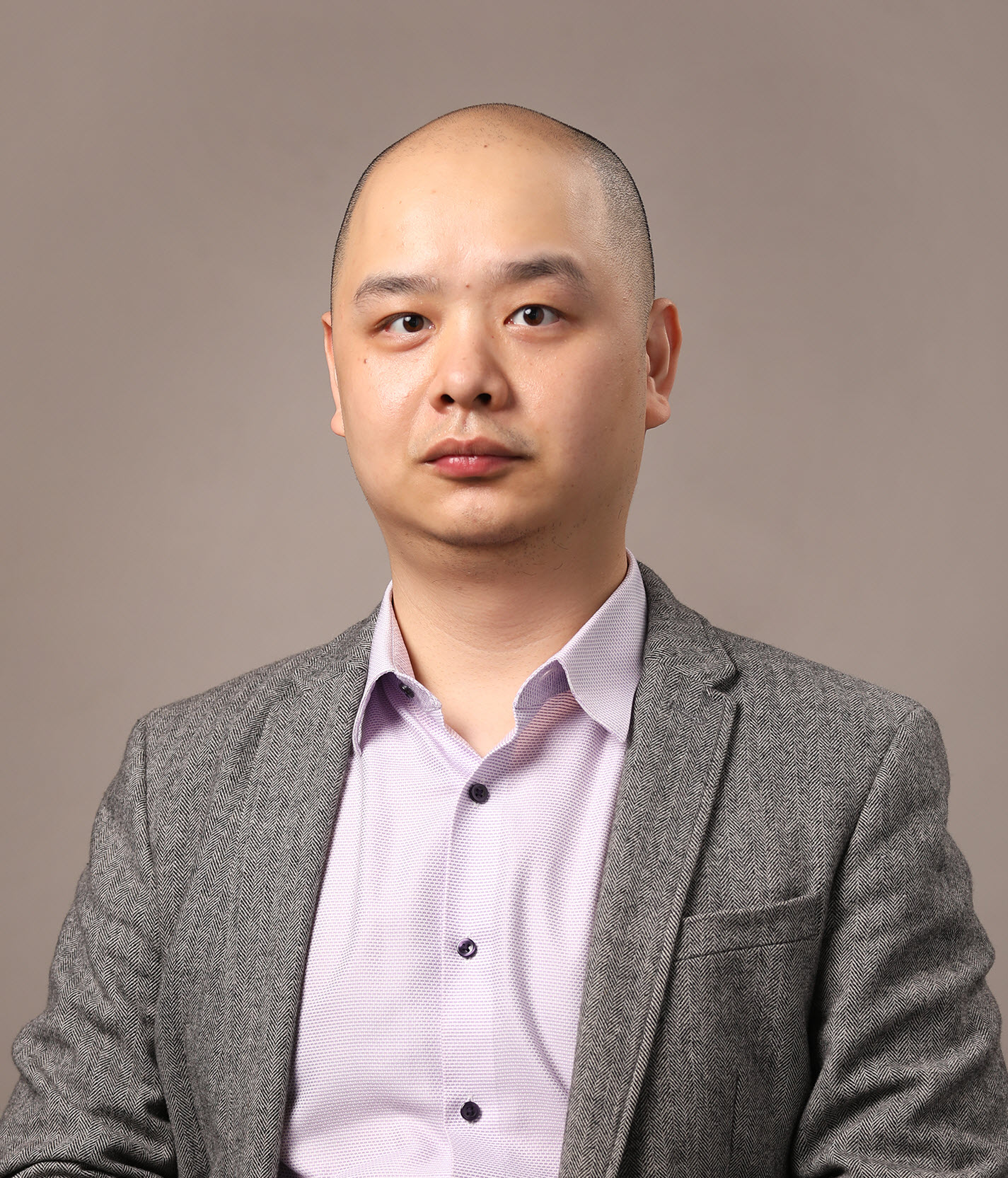}}]{Haoran Xie (Senior member, IEEE)} is Associate Professor at the Department of Computing and Decision Sciences, Lingnan University, Hong Kong SAR. His research interests include artificial intelligence, big data, and educational technology. He has published 350 research publications, including 192 journal articles such as IEEE TPAMI, IEEE TKDE, IEEE TAFFC, IEEE TCVST, and so on. He has been selected as The World Top 2\% Scientists by Stanford University.
\end{IEEEbiography}
\vskip -3\baselineskip plus -1fil
\begin{IEEEbiography}[{\includegraphics[width=1in,height=1.25in,clip,keepaspectratio]{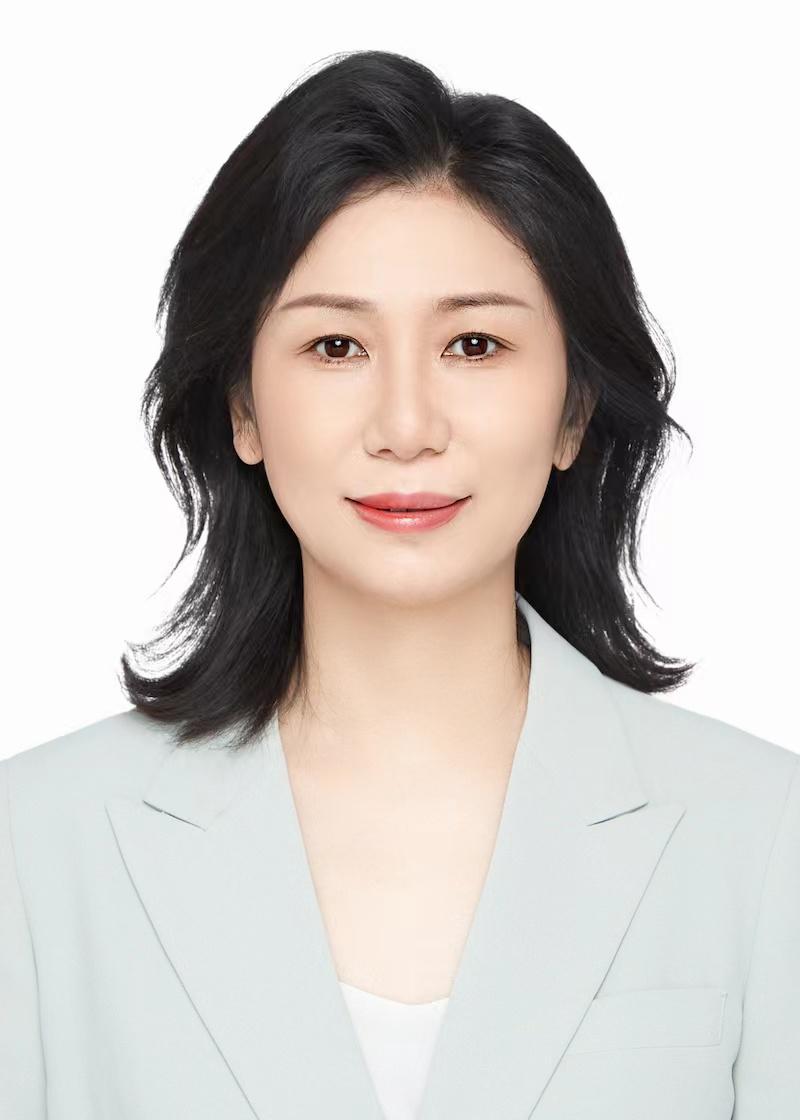}}]{Lin Li} is Professor at the School of Computer Science and Artificial Intelligence, Wuhan University of Technology. Her research interest covers Information Retrieval, Recommender Systems, Data Mining, Multimedia Computing, and Natural Language Processing. Her 100+ publications on IJCAI, AAAI,  WWW, ICDM, ICMR, CIKM, DASFAA, TOIS, TSC and TOIT have received 2100+ citations on Google Scholar. She received her PhD from University of Tokyo, Japan in 2009.
\end{IEEEbiography}
\vskip -3\baselineskip plus -1fil
\begin{IEEEbiography}[{\includegraphics[width=1in,height=1.25in, clip,keepaspectratio]{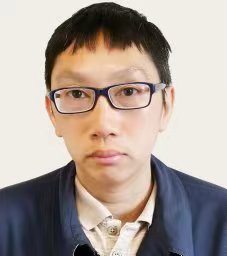}}]{Xiaofeng Zhu (Senior Member, IEEE)} received his Ph.D. degree in Computer Science from The University of Queensland, Brisbane, QLD, Australia, in 2013. He is currently a Professor in the School of Computer Science and Engineering at the University of Electronic Science and Technology of China, Chengdu, China.  His research interests include machine learning, graph learning, and medical imaging analysis.
\end{IEEEbiography}
\vskip -3\baselineskip plus -1fil
\begin{IEEEbiography}[{\includegraphics[width=1in,height=1.25in, clip,keepaspectratio]{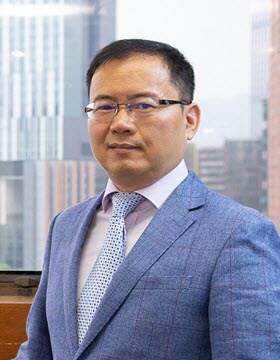}}]{Qing Li (Fellow, IEEE)}
is currently a Chair Professor (Data Science) and the Head of the Department of Computing, the Hong Kong Polytechnic University. He is currently a Fellow of IEEE and IET/IEE, and a Distinghished Member of CCF (Chinese Computer Federation). He is the chairperson of the Hong Kong Web Society, and a Distinguished Member of CCF. He has been a Steering Committee member of DASFAA, ER, ICWL, UMEDIA, and WISE Society.

\end{IEEEbiography}


\newpage

\vfill{40cm}

\includepdf[pages=-]{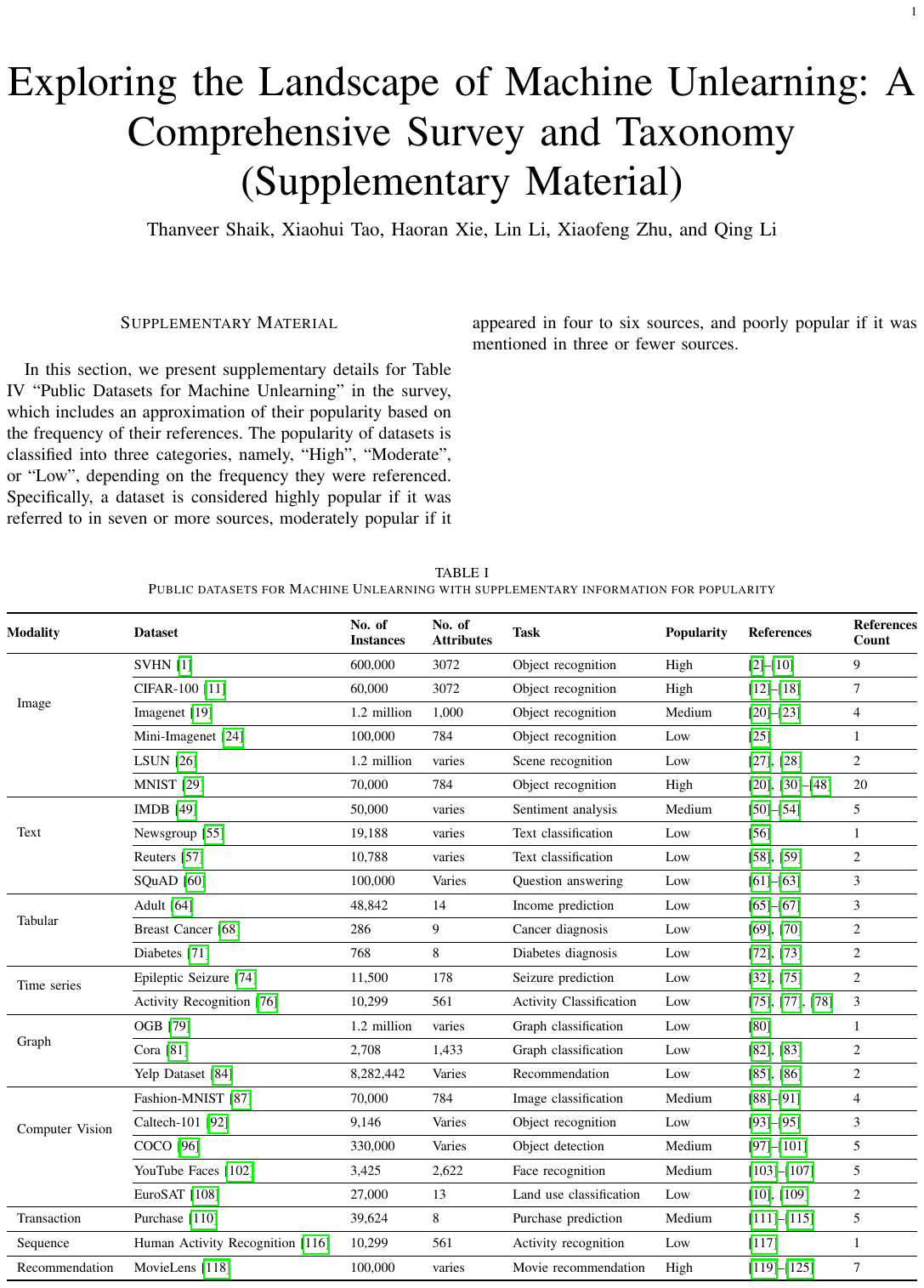}
\end{document}